\documentclass[runningheads]{llncs}
\usepackage{graphicx}
\usepackage{amsmath,amssymb} 
\usepackage{color}

\usepackage{subfigure}
\usepackage{amsfonts}
\usepackage{booktabs}
\usepackage{tabu}
\usepackage{siunitx}
\usepackage{enumitem}
\usepackage{multirow}
\usepackage{capt-of}

\newcommand{\etal}{\textit{et al.}}

\begin{document}
\title{SAN: Learning Relationship between \\ Convolutional Features \\for Multi-Scale Object Detection} 

\titlerunning{Scale Aware Network for Object Detection}
%
\author{Yonghyun Kim\inst{1}\orcidID{0000-0003-0038-7850} \and
	Bong-Nam Kang\inst{2}\orcidID{0000-0002-6818-7532} \and
	Daijin Kim\inst{1}\orcidID{0000-0002-8046-8521}}
\index{Yonghyun, Kim}
\index{Bong-Nam, Kang}
\index{Daijin, Kim}
%
\authorrunning{Y. Kim~\etal}
%

\institute{Department of Computer Science and Engineering, POSTECH, Korea\and
	Department of Creative IT Engineering, POSTECH, Korea\\
	\email{ \{gkyh0805,bnkang,dkim\}@postech.ac.kr}
}
\maketitle              
\begin{abstract}
	Most of the recent successful methods in accurate object detection build on the convolutional neural networks (CNN).
	However, due to the lack of scale normalization in CNN-based detection methods, the activated channels in the feature space can be completely different according to a scale and this difference makes it hard for the classifier to learn samples. 
	We propose a Scale Aware Network (SAN) that maps the convolutional features from the different scales onto a scale-invariant subspace to make CNN-based detection methods more robust to the scale variation, and also construct a unique learning method which considers purely the relationship between channels without the spatial information for the efficient learning of SAN.
	To show the validity of our method, we visualize how convolutional features change according to the scale through a channel activation matrix and experimentally show that SAN reduces the feature differences in the scale space.
	We evaluate our method on VOC PASCAL and MS COCO dataset.
	We demonstrate SAN by conducting several experiments on structures and parameters.	
	The proposed SAN can be generally applied to many CNN-based detection methods to enhance the detection accuracy with a slight increase in the computing time. 
\end{abstract}
\section{Introduction}

Accurate and efficient detection of multi-scale objects is an important goal in object detection.
Multi-scale objects are detected by a single detector~\cite{dalal2005histograms,dollar2014fast,dollar2009integral,viola2004robust} that uses an image pyramid with scale normalization,
or by a multi-scale detector~\cite{benenson2012pedestrian} that uses a separate detector for each of several scales.
However, detection methods that are based on a convolutional neural network (CNN) can detect multi-scale objects by pooling regions of interest (RoIs)~\cite{girshick2015fast,li2016r,ren2015faster} to extract convolutional features of the same size in RoIs of different sizes,
or by learning grid cells~\cite{redmon2016you,redmon2017yolo9000} that represent the surrounding area, or by assigning different scales according to the level of the feature map~\cite{lin2017feature,liu2016ssd}, without scale normalization.

The process of scale normalization can cause differences in feature space between samples if they have different resolutions.
By mapping samples from different resolutions to a common subspace~\cite{yan2013robust} or by calibrating the gradient features from different resolutions to the gradient features at the reference resolution~\cite{detector2017icip},
the variation between samples is reduced and the detection accuracy is improved.

\begin{figure}[t]
	\includegraphics[height=3.5cm]{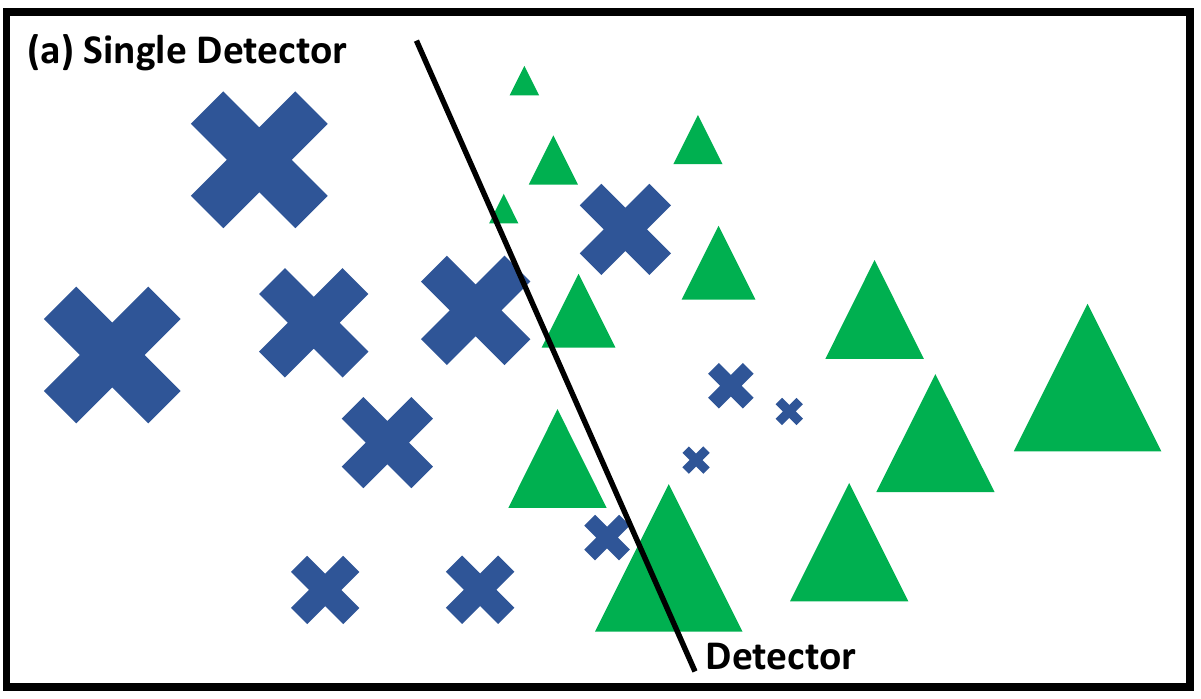}
	\includegraphics[height=3.5cm]{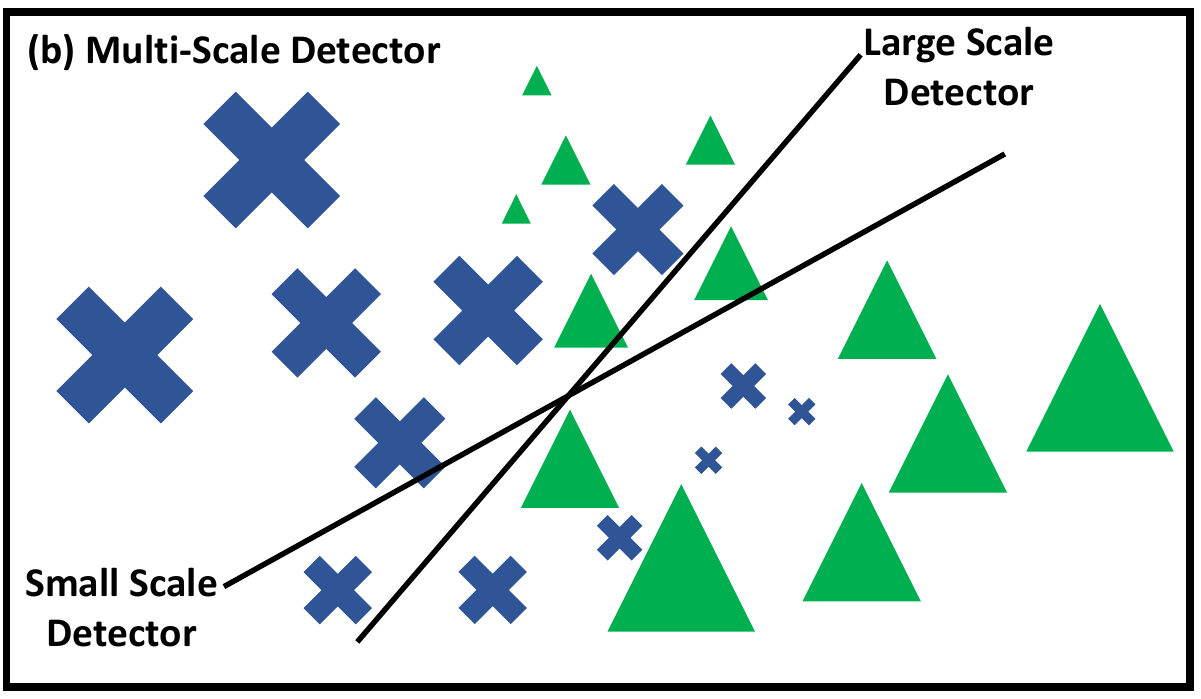}\\
	\includegraphics[height=3.5cm]{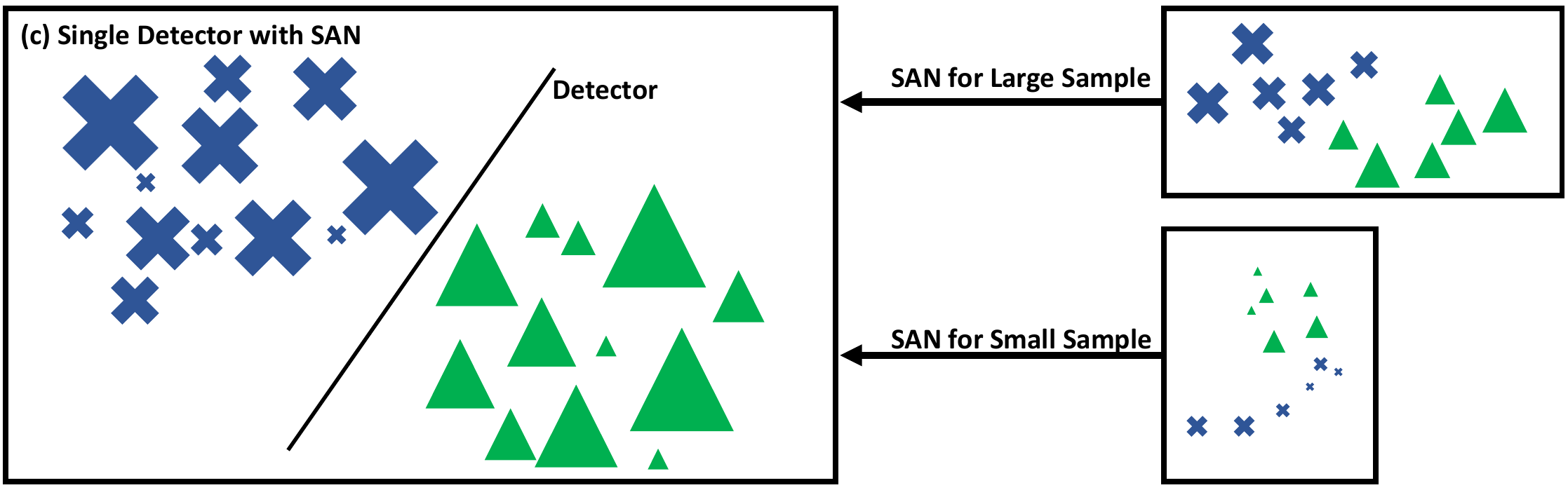}
	\caption{Different strategies for multi-scale object detection. The blue cross and green triangular marks represent background and object samples, respectively, and the size of the mark is proportional to the size of the sample}
	\centering
	\label{fig:SAN_CONCEPTS}
\end{figure}

However, CNN-based detection methods generally do not perform scale normalization,
so new differences arise due to scale rather than to resolution.
CNN-based methods that use RoI pooling or grid cells may represent the sizes of cell differently according to the size of RoIs,
so this scale variation can lead to extraction of completely different shapes of features rather than to a small difference in the resolution variation.
Several strategies can be used to detect multi-scale objects (Fig.~\ref{fig:SAN_CONCEPTS}).
A single detector strategy (Fig.~\ref{fig:SAN_CONCEPTS}a) is used in many detection methods; it has a simple structure and can learn a single classifier from the whole set of training samples,
but it may not consider the scale space sufficiently.
The multi-scale detector strategy (Fig.~\ref{fig:SAN_CONCEPTS}b) learns multiple detectors for different scale spaces,
but may have difficulty in learning the classifications if the scale space for each detector has too few samples.

We propose a Scale Aware Network (SAN) that maps the convolutional features from the different scales onto a scale-invariant subspace to learn a single classifier with consideration of the scale space (Fig.~\ref{fig:SAN_CONCEPTS}c).
SAN learns the relationships between convolutional features in the scale space to reduce the feature differences and improves the detection accuracy.
We study the effect of the scale difference in CNN by using a channel activation matrix that represents the relationship between scale change and channel activation,
then design a structure for SAN and a unique learning method based on channel routing mechanism that considers the relationship between channels without the spatial information.

We make three main contributions: 
\begin{itemize}[leftmargin=+.4in,label=$\bullet$]
	\item We develop SAN that maps the convolutional features from the different scales onto a scale-invariant subspace.
	We study the relationship between scale change and channel activation and, based on this,
	we design a unique learning method which considers purely the relationship between channels without the spatial information.
	The proposed SAN reduces the feature differences in the scale space and improves the detection accuracy.
	
	\item We empirically demonstrate SAN for object detection by conducting several experiments on structures and parameters for SAN.
	We visualize how convolutional features change according to the scale through a channel activation matrix and 
	experimentally prove that SAN reduces the feature differences in the scale space.

	\item The proposed SAN essentially improves the quality of convolutional features in the scale space,
	thus it can be generally applied to many CNN-based detection methods to enhance the detection accuracy with a slight increase in the computing time. 
\end{itemize}

This paper is organized as follow.
We review related works in Section 2.
We discuss the effect of the scale difference in CNN and present the proposed SAN and the training mechanism for SAN in Section 3.
We show experimental results for object detection and empirically demonstrate SAN in Section 4.
We conclude in Section 5.

\section{Related Works}

\noindent
\textbf{Multi-Scale Detection.} 
The image pyramid~\cite{adelson1984pyramid,gonzalez2009digital}, which is one of the most popular approaches to multi-scale detection, has been applied to many applications such as pedestrian detection~\cite{dalal2005histograms,ding2012contextual,dollar2014fast,dollar2009integral}, human pose estimation~\cite{yang2013articulated}, and object detection~\cite{felzenszwalb2010object}. 
Because the image pyramid is constructed by resampling a given image to multiple scales, the differences in image statistics can occur depending on the degree of resampling.
Several researchers studied natural image statistics and the relationship between two resampled images~\cite{field1987relations,ruderman1994statistics,ruderman1994statistics1}.
The difference in resolution caused by resampling degrades the detection accuracy.
The variation between samples can be reduced by mapping samples from different resolutions onto a common subspace~\cite{yan2013robust} or by calibrating the gradient features from different resolutions to the gradient features at the reference resolution~\cite{detector2017icip}, and the reduced variance makes  it easy for the classifier to learn samples.
In this work, we discuss the feature difference in CNN caused by the scale variance and make CNN-based detection methods more robust to the scale variation.
\\

\noindent
\textbf{Detection Networks.} 
The development of deep neural networks has achieved tremendous performance improvements in the field of object detection.
Especially, Faster R-CNN~\cite{ren2015faster}, which is one of the representative object detection algorithms,
generates candidate proposals using a region proposal network and classifies the proposals to the background and foreground classes using RoI pooling.
Region-based fully convolutional networks (R-FCN)~\cite{li2016r} improved speed by designing the structure of networks as fully convolutional by excluding RoI-wise sub-networks.
PSRoI pooling in R-FCN solved the translation dilemma without a deep RoI-wise sub-network.
R-FCN achieved the same detection performance as Faster R-CNN at faster speed.
Deformable R-FCN~\cite{dai17dcn} suggests deformable convolution and RoI pooling, which are a generalization of atrous convolution.
SAN is trained using the relationship between convolutional features extracted by RoI pooling in the scale variation.
We improve the accuracy of object detection by applying the proposed SAN between the last layer of ResNet-101 and the $1\times1$ convolutional layers for detection of R-FCN and Deformable R-FCN.

\noindent
\textbf{Residual Network.}
The residual network~\cite{he2016deep}, one of the most widely used backbone networks in recent years, was proposed to solve the problem that learning becomes difficult as the network becomes deeper.
The residual learning prevents the deeper networks from having a higher training error than the shallower networks by adding shortcut connections that are identity mapping.
ResNet-101 is constructed using this residual learning with 101 layers.
In this work, the detection network is based on a fully convolutional network that excludes the average pooling, 1000-d fully connected and softmax layers from ResNet-101.
We modify a stride of the last convolution block \textit{res5} from 1 to 2 for doubling the receptive fields in detection.
The dilation is changed from 1 to 2 for all $3\times3$ convolution to compensate this modification.

\noindent
\textbf{Resolution Aware Detection Model.}
Many existing detectors find objects of various sizes by sliding a detection model on an image pyramid.
The window used as the input of the detector is normalized to a pre-defined size, but there is a resolution difference in the resampling process.
A resolution aware detection model~\cite{yan2013robust} reduces the resolution difference by considering the relationships between the samples obtained at different resolutions, and trains a detection model and a resolution aware transformation to map features from different resolutions to a common subspace.
Since the detector learns only the samples on the common subspace, the variation between samples can be reduced and higher detection accuracy can be achieved.
The concepts, which reduce the variance of samples, are widely used to improve the performance of a classifier~\cite{geman1992neural,james2013introduction}.
A typical way of reducing the variance is partitioning samples by pose~\cite{huang2005vector,jones2003fast,wu2004fast}, rotation~\cite{yang2013articulated} or resolution~\cite{park2010multiresolution,yan2013robust}.
They reduced the coverage of a classifier by reducing the variance of samples, and the decreased coverage improved the performance of a classifier.
In a similar manner, the proposed SAN suggests a method of mapping to a scale-invariant subspace in consideration of the relationship between different scales, not the resolution, to overcome the lack of scale normalization in CNN.
\\

\begin{figure}[t]
	\centering
	\subfigure[with Scale Normalziation]{\includegraphics[height=5.7cm]{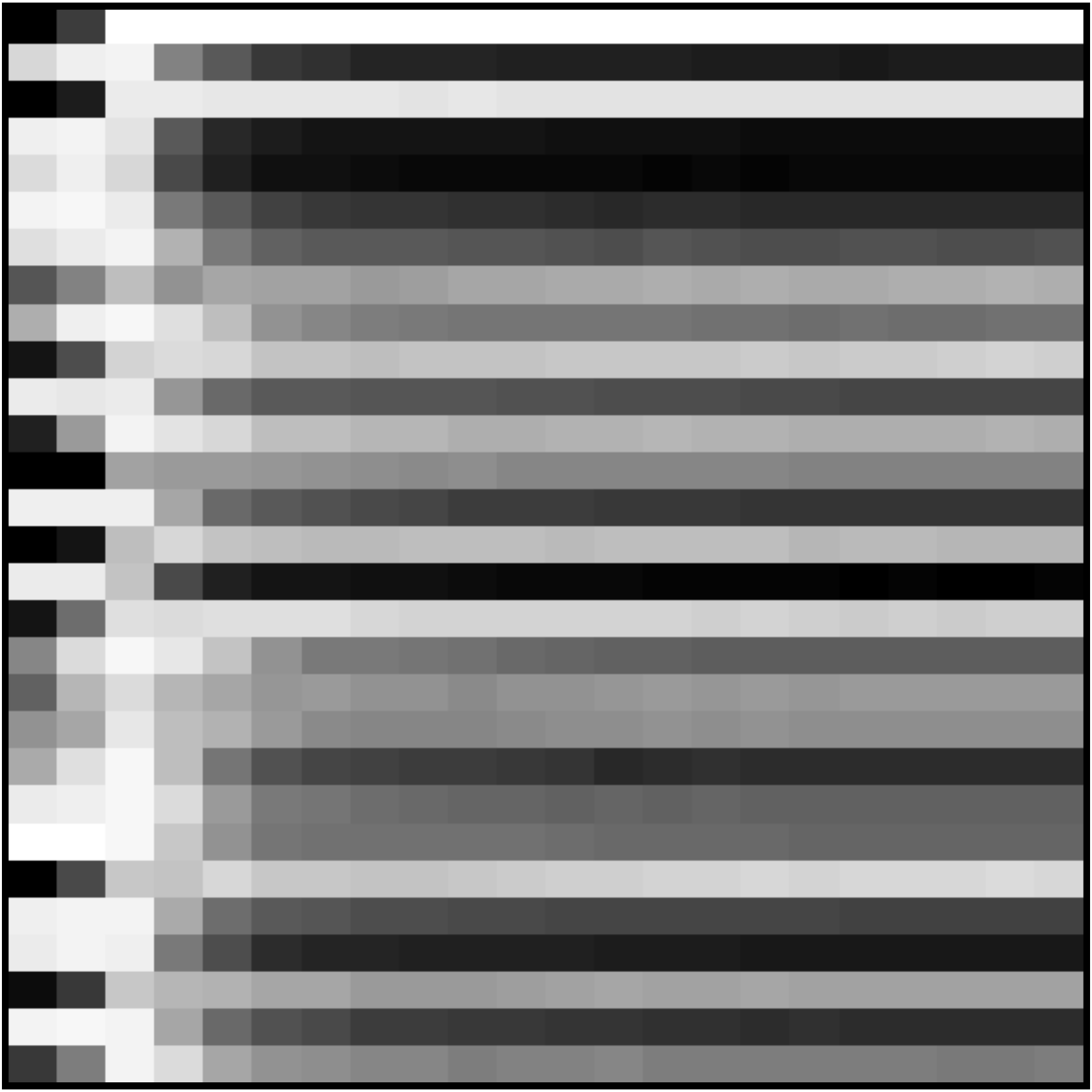}}
	\subfigure[without Scale Normalziation]{\includegraphics[height=5.7cm]{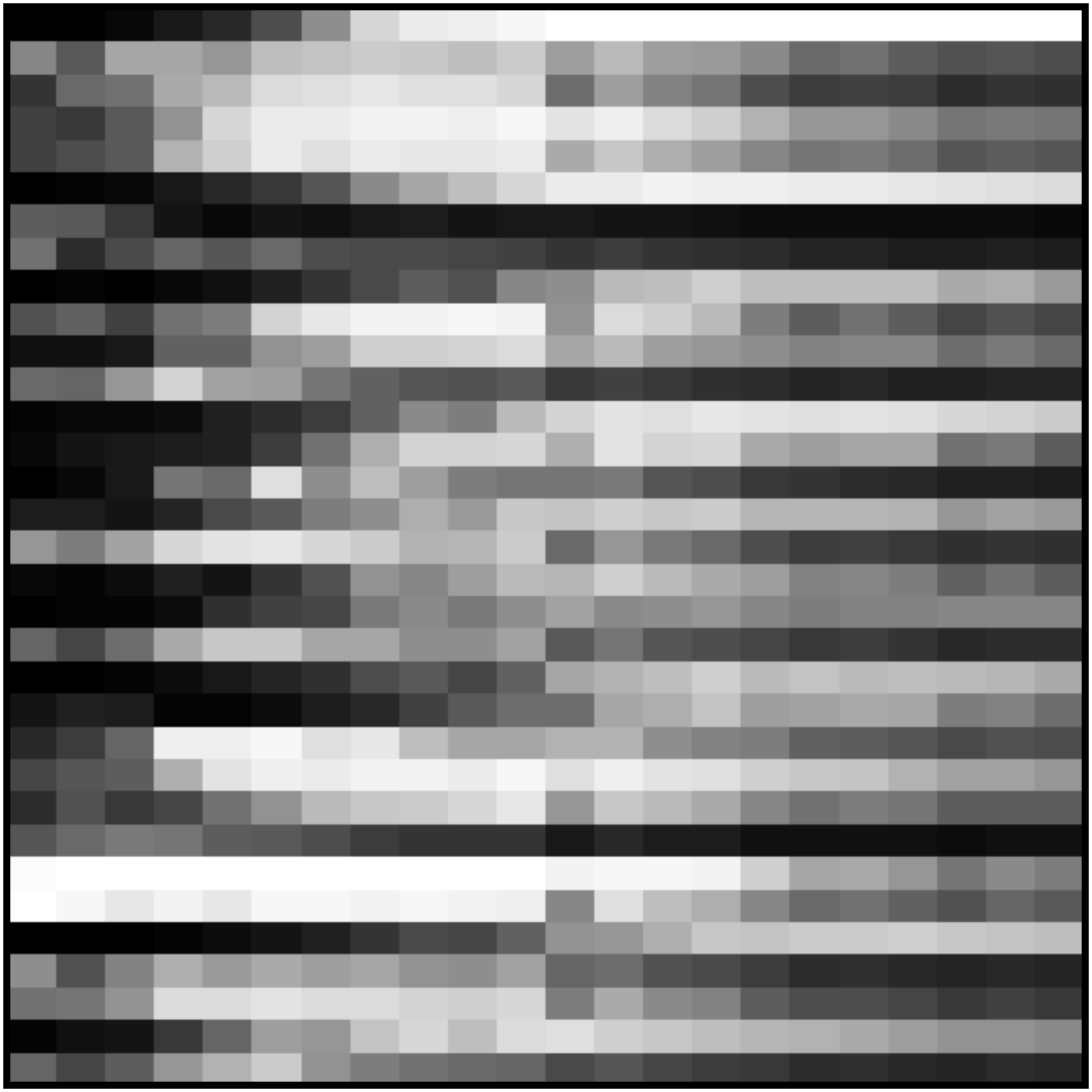}}
	\caption{
		The channel activation matrix for the scale variation shows the comparison of RoI pooling with and without scale normalization.
		The x-axis represents the scale of an image and the y-axis represents the channel index.
		The closer the block is to white, the more the activation of the corresponding channel in the block
	}
	\label{fig:SAN_MOTIVE}
\end{figure}

\section{Scale Aware Network}

\textbf{Overview.}
We propose a Scale Aware Network (SAN) that maps the convolutional features from the different scales onto a scale-invariant subspace to make CNN-based detection methods more robust to the scale variation.
CNN-based detectors have higher detection accuracy than the conventional detectors without any scale normalization.
The conventional detectors, which mainly use hand-crafted features, detect multi-scale objects by classifying the scale normalized patches.
The features obtained from the scale normalized patches have a small differences in the resolution variation caused by resizing, but the sizes of the objects and parts in the images remain unchanged.
However, the convolutional features can completely change the activated channels rather than vary only slightly due to the lack of scale normalization.
\\

\noindent
\textbf{Channel Activation Matrix.} 
High dimensionality complicates the task of visually observing how convolutional features change according to the scale.
The channel activation matrix (CAM) shows how convolutional features are affected by the scale variation by comparing values only of the channels that are primarily activated.
Comparison of CAM for RoI Pooling with and without scale normalization (Fig.~\ref{fig:SAN_MOTIVE}) shows the difference of the channel-wise output of the last residual block \textit{res5c} in ResNet-101 according to the scale variation.
We calculate CAM by redundantly extracting 10 channels that show large activation for resized images from $8\times8$ to $448\times448$.
CAM for convolutional features with scale normalization (Fig.~\ref{fig:SAN_MOTIVE}a) shows uniform channel activation in most of the scale space except that the resolution is severely impaired when the scale is too small.
In contrast, CAM for convolutional features without scale normalization (Fig.~\ref{fig:SAN_MOTIVE}b) shows non-uniform channel activation across the scale space, and its value varies with the change in scale.
Due to non-uniformity in the scale space, the scale variation can degrade the detection accuracy.
\\

\begin{figure}[t]
	\includegraphics[width=12cm]{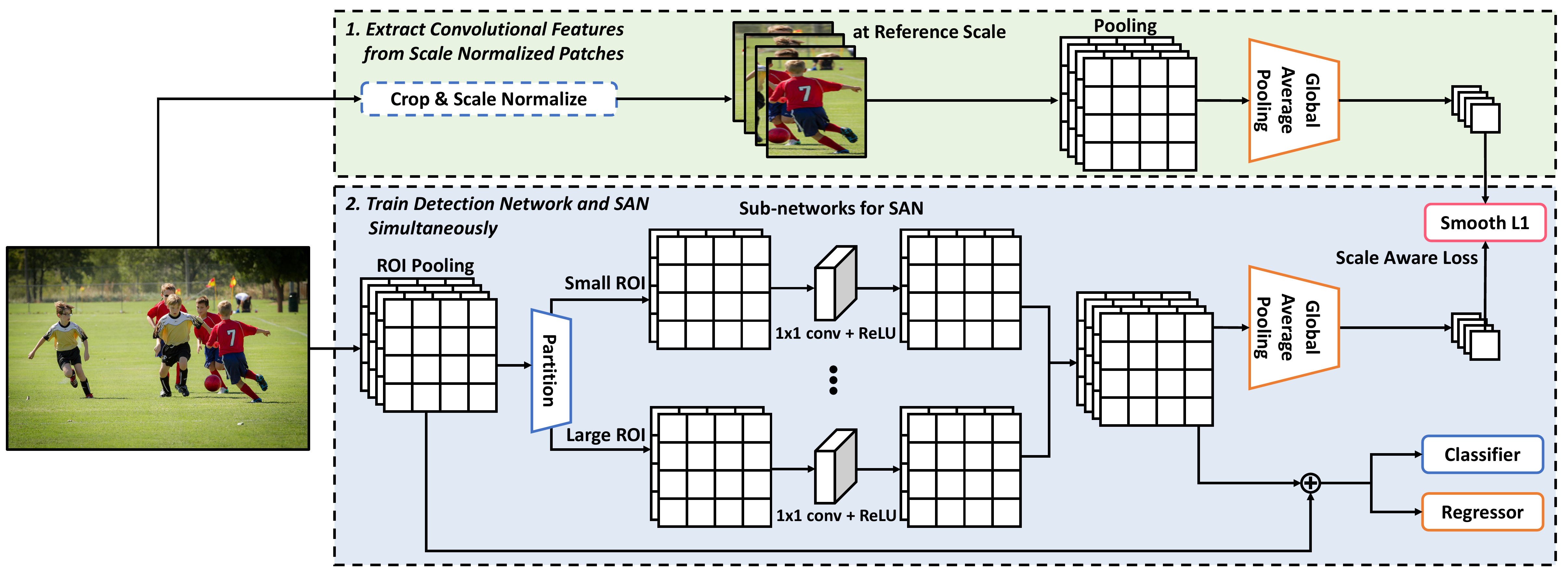}
	\caption{Architecture of SAN.
		In learning, an additional stage is required for SAN to extract convolutional features from scale normalized patches.
		The detection network and SAN is trained using the extracted features, simultaneously.
		In inference, SAN simulates the scale normalization without it
	}
	\centering
	\label{fig:SAN_ARCH}
\end{figure}

\noindent
\textbf{Architecture.}
SAN consists of several sub-networks corresponding to the different sizes of RoI.
By exploiting these sub-networks, SAN learns the relationship between convolutional features obtained from the scale-normalized patch and RoI pooling of the input image~(Fig.~\ref{fig:SAN_ARCH}).
Each sub-network consists of a $1\times1$ convolution of which the number of channels equals to the input feature and the following ReLU. 
We partition RoIs in a mini-batch into three intervals of size at the reference scale experimentally: 
$(0^2,160^2]$, $(160^2,288^2)$, $[288^2,\infty)$ at $224^2$ for VOC Pascal 
and
$(0^2,64^2]$, $(64^2,192^2)$, $[192^2,\infty)$ at $128^2$ for MS COCO. 
The partitioned features are corrected by using the corresponding sub-network of SAN, then merged into one mini-batch.
The detector uses element-wise sum of the original feature and the SAN feature to enrich the feature representation, just as the low-level features in FPN~\cite{lin2017feature}.
SAN can be adapted to many other types of CNN-based detection framework with a simple network extension.
\\

\begin{figure}[t]
	\includegraphics[width=12cm]{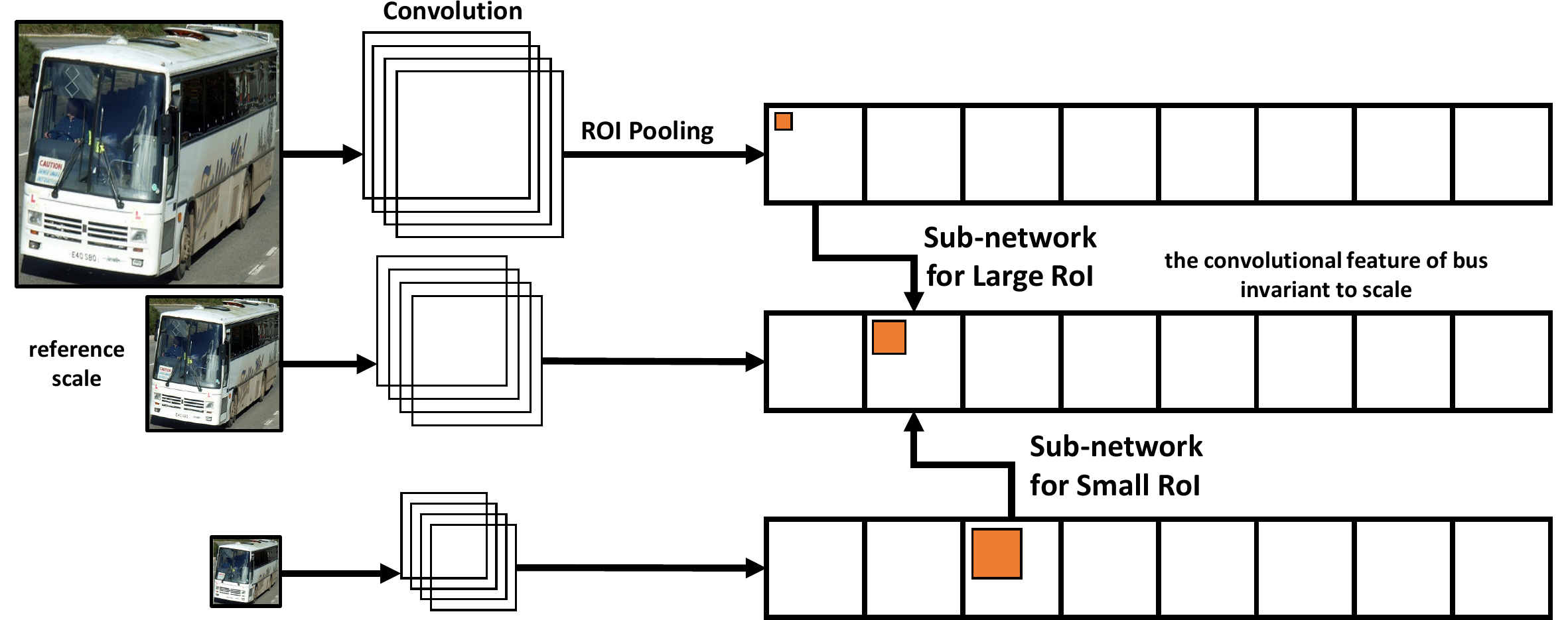}
	\caption{A mechanism of channel routing.
		The smaller the size of the image, the lower the resolution in the feature space.
		Therefore, several channels can imply the same meaning according to the scale.
		Because of the difference in spatiality caused by the difference in receptive fields, learning should be done with the concept of channel routing without the spatial information}
	\centering
	\label{fig:SAN_CHANNEL_ROUTING}
\end{figure}

\noindent
\textbf{Channel Routing Mechanism.} 
The different receptive fields at various scales cause the discordance of the spatial information.
The discordance makes SAN difficult to learn the expected scale normalization precisely. 
Thus, we use a global average pooling to learn only the information of channels by excluding the discordance of the spatial information.
We interpret this learning method as a concept of routing: SAN transforms a channel activation at the specific scale to a channel activation at the reference scale~(Fig.~\ref{fig:SAN_CHANNEL_ROUTING}).
\\

\noindent
\textbf{Loss Functions.} 
The entire detection framework is divided into two parts according to the influence of loss: 
(1) SAN: SAN trained with a combination of classification, box regression, and scale-aware losses,
and (2) Network without SAN: networks excluding SAN trained using only classification and box regression loss~\cite{girshick2015fast,ren2015faster,li2016r}.
Scale-aware loss is designed to reduce the difference between features over scales, so it can cause an error in detection when applied to the entire network.
In the case of SAN, both the scale invariance and the detection accuracy must be considered, so all three losses are applied. 
The multi-task loss for SAN is defined as:

\begin{equation}
\mathcal{L}(p,u,t^u,v,r,\widetilde{r}) = 
L_{cls} (p,u) 
+ [u \geq 1] L_{reg}(t^u,v)
+ L_{san}(r,\widetilde{r}).
\end{equation}

Here, $p$ is a discrete probability distribution over $K+1$ categories and $u$ is a ground-truth class.
$t^k$ is a tuple of bounding-box regression for each of the $K$ classes, indexed by $k$, and $v$ is a tuple of ground-truth bounding-box regression.
$r$ is a channel-wise convolutional feature extracted from RoI and $\widetilde{r}$ is a channel-wise convolutional feature for a scale-normalized patch from RoI.
The classification loss,
$L_{cls} (p,u)=-\log p_u$,
is logarithmic loss for ground-truth class $u$.
The regression loss,
$L_{reg}(t^u,v)=\sum _{i \in \{ x,y,w,h \} } {\mathbf{smooth}_{L_1}} \left[ t_i^u-v_i \right]$,
measures the difference between $t^u$ and $v$ using the robust $L_1$ function~\cite{girshick2015fast}.
The scale-aware loss $L_{san}$ represents the difference between $r$ and $\widetilde{r}$ using the robust $L_1$ function, and defined as: 
\begin{equation}
L_{san}(r,\widetilde{r}) = \sum _ {c \in C} {\mathbf{smooth}_{L_1}} \left[ r_c-\widetilde{r}_c \right].
\end{equation}

In this work, we extract the convolutional features~$\widetilde{r}$ for the scale-normalized patches from only 16 randomly selected RoIs in a mini-batch due to the computing time.

\begin{figure}[t]
	\includegraphics[width=12cm]{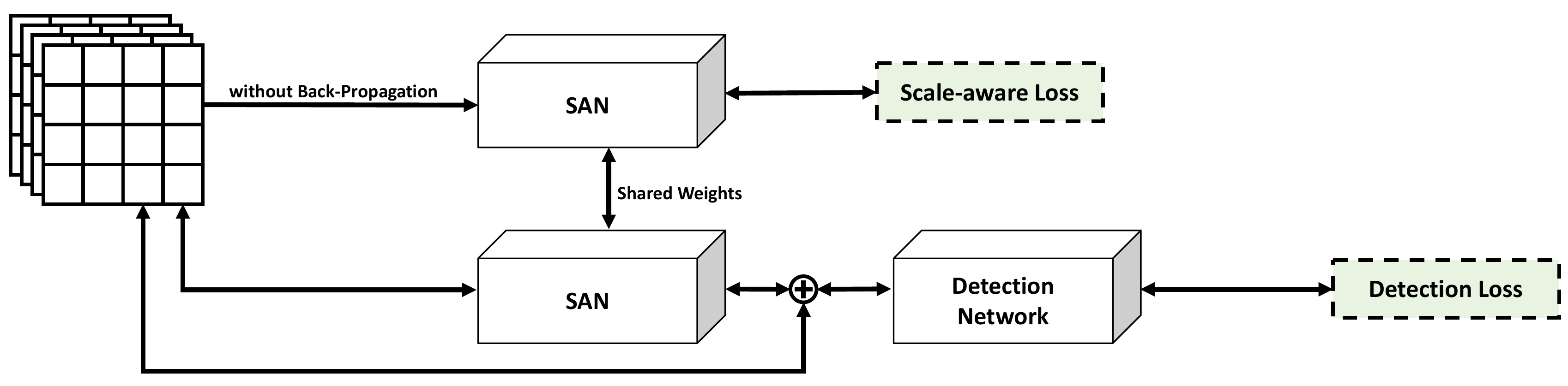}
	\caption{A learning trick for SAN.
		The entire network uses two SANs sharing the weight as the siamese architecture to control the influence of losses
	}
	\centering
	\label{fig:SIAMESE}
\end{figure}

\section{Experiments}

\noindent
\textbf{Structure.}
We experimentally demonstrate the effectiveness of SAN based on R-FCN~\cite{li2016r}.
In this work, the backbone of R-FCN is ResNet-101~\cite{he2016deep}, which consists of 100 convolutional layers followed by global average pooling and a 1000 class fully-connected layer.
We leave only the convolutional layers to compute feature maps by removing the average pooling layer and the fully-connected layer.
We attach a $1\times1$ convolutional layer as a feature extraction layer, which consists of 1024 channels and is initialized from a gaussian distribution, to the end of the last residual block in ResNet-101.
We apply a convolutional layers for classification and box regression, and extract a $7\times7$ score and regression map for given RoI using PSRoI pooling.
Then, the probability of classes and the bounding boxes for the corresponding RoI are predicted through average voting.
SAN, which is applied between the feature extraction layer and the detection layer, consists of several sub-networks corresponding to the different sizes of RoI and each sub-network consists of a $1\times1$ convolution of which the number of channels equals to the input feature and the following ReLU.
\\

\noindent\textbf{Learning.}
The detection network is trained by stochastic gradient descent (SGD) with the online hard example mining (OHEM)~\cite{shrivastava2016training}.
We use the pre-trained ResNet-101 model on ImageNet~\cite{russakovsky2015imagenet}.
We train the network for
29k iterations with a learning rate of $10^{-3}$ dividing it by 10 at 20k iterations, a weight decay of $0.0005$, and a momentum of $0.9$ for VOC PASCAL 
and
240k iterations with a learning rate of $10^{-3}$ dividing it by 10 at 160k iterations, a weight decay of $0.0005$, and a momentum of $0.9$ for MS COCO.
A mini-batch consists of 2 images, which are resized such that its shorter side of image is 600 pixels.
We use 300 proposals per image for training and testing.
\\

\noindent
\textbf{Learning Trick for SAN.}
We divide the entire detection framework into two parts according to the influence of loss to exclude the effect of scale-aware loss on the detection framework.
However, since it is difficult to separate only the loss corresponding to SAN from the already aggregated loss, we use a learning trick based on the siamese architecture~\cite{chopra2005learning}.
We configure the siamese architecture of two SAN with shared weights for scale-aware and detection loss, respectively (Fig.~\ref{fig:SIAMESE}).
By not propagating the error from the siamese network for scale-aware loss, we prevent the performance degradation caused by scale-aware loss and make SAN consider both scale invariance and detection.
Without this learning trick, the extension with SAN can reduces the detection accuracy.
\\

\newcolumntype{P}[1]{>{\centering\arraybackslash}p{#1}}
\begin{table}[t!]
	\caption{Comparison on VOC PASCAL and MS COCO. 
		$\textsl{SAN}_0$ stands for SAN without scale-aware loss
		and $\textsl{SAN}$ stands for the full extension of SAN with scale-aware loss}
	\begin{center}
		\scalebox{0.95}{
			\begin{tabular}{l|P{2.2cm}|P{1.2cm}|P{1.2cm}|P{1.2cm}|P{1.2cm}|P{1.2cm}}
				\toprule 
				& \scalebox{0.9}{VOC 07+12/07}   & \multicolumn{5}{c}{\scalebox{0.9}{MS COCO \texttt{trainval35k}/\texttt{minival}}}  \\
				& \scalebox{0.9}{$\mathrm{mAP}$} & \scalebox{0.9}{$\mathrm{mAP}$} & \scalebox{0.9}{$\mathrm{mAP_{@0.5}}$} & 
				\scalebox{0.9}{$\mathrm{mAP_{@S}}$} & \scalebox{0.9}{$\mathrm{mAP_{@M}}$} & \scalebox{0.9}{$\mathrm{mAP_{@L}}$}  \\
				\midrule 
				Faster RCNN~\cite{ren2015faster}						&  79.3 & 30.3 & 52.1 &  9.9 & 32.2 & 47.4 \\
				FPN~\cite{lin2017feature}		&  - 	& 37.8 & 60.8 & 22.0 & 41.5 & 49.8 \\
				R-FCN~\cite{li2016r}							&  79.4 & 35.3 & 58.7 & 16.4 & 39.5 & 53.8 \\
				Deformable R-FCN~\cite{dai17dcn}&  82.1 & 41.2 & 62.9 & 19.4 & 46.6 & 61.2 \\
				\midrule 
				R-FCN-$\textsl{SAN}_0$ 					&  80.1 & - & - & - & - & - \\
				Deformable R-FCN-$\textsl{SAN}_0$ 		&  82.4 & - & - & - & - & - \\
				\midrule 
				Faster RCNN-\textsl{SAN}			&  79.9 & - & - & - & - & - \\
				R-FCN-\textsl{SAN} 					&  80.6 & 36.3 & 59.6 & 16.7 & 40.5 & 55.5 \\
				Deformable R-FCN-\textsl{SAN}		&  82.8 & 43.3 & 65.1 & 21.0 & 48.8 & 64.6 \\
				\bottomrule 
			\end{tabular}
		}
	\end{center}
	\label{tab:tablec}
\end{table}

\noindent
\textbf{Experiments on PASCAL VOC.}
We evaluate the proposed SAN on PASCAL VOC dataset~\cite{everingham2010pascal} that has 20 object categories.
We train the models on the union set of VOC 2007 \texttt{trainval} and VOC 2012 \texttt{trainval} (16k images), and evaluate on VOC 2007 \texttt{test} set (5k images).
We attach SAN to baseline under these conditions: 
the reference scale of $224^2$, 
the three partitions of $(0^2,160^2]$, $(160^2,288^2)$, $[288^2,\infty)$. 
SAN improves the mean Average Precision (mAP) by 1.2 points with a slight increase from 120ms to 130ms in the computing time, over a single-scale baseline of R-FCN on ResNet-101.
\\

\noindent
\textbf{Experiments on MS COCO.}
We evaluate the proposed SAN on MS COCO dataset~\cite{lin2014microsoft} that has 80 object categories.
We train the models on the union set of 80k training set and a 35k subset of validation set (\texttt{trainval35k}),
and evaluate on a 5k subset of validation set~(\texttt{minival}).
We attach SAN to baseline under these conditions: 
the reference scale of $128^2$, 
the three partitions of $(0^2,64^2]$, $(64^2,192^2)$, $[192^2,\infty)$. 
SAN improves the COCO-style mAP, which is the average AP across thresholds of IoU from 0.5 to 0.95 with an interval of 0.05, by 2.1 points over a single-scale baseline of Deformable R-FCN on ResNet-101.
\\

\begin{figure}[t]
	\includegraphics[width=12cm]{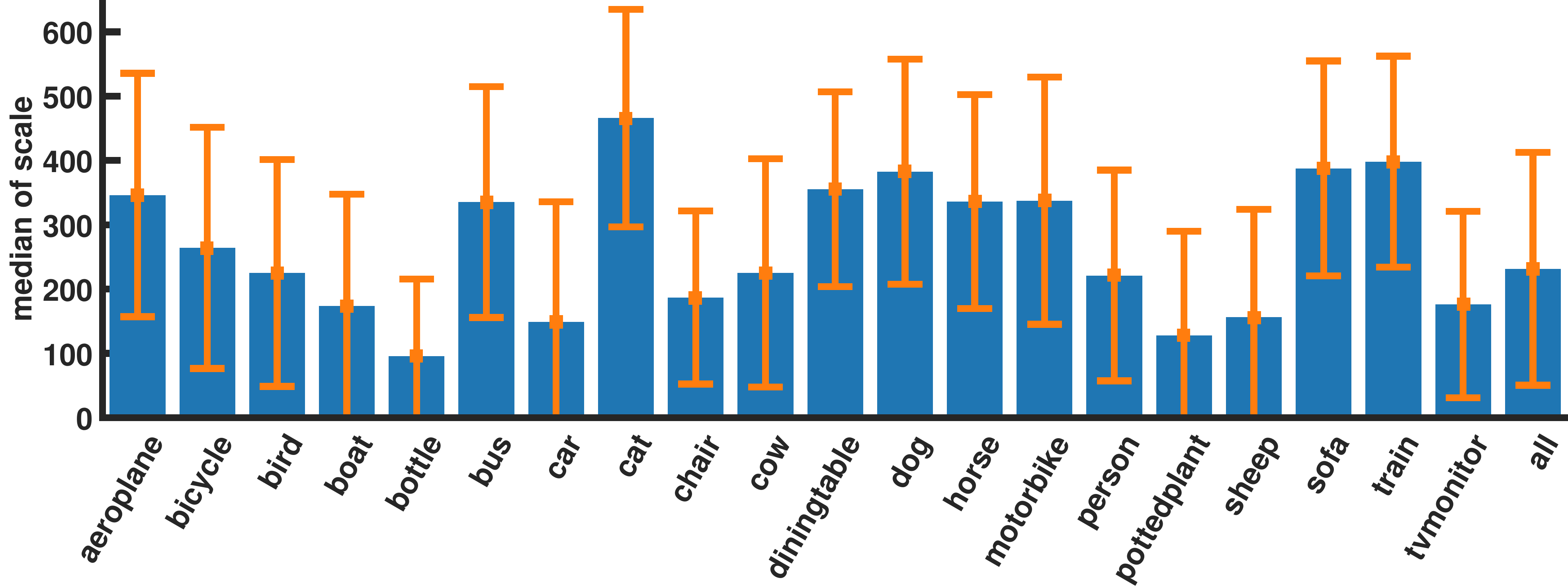}
	\caption{The medians and standard deviations of scales according to the classes}
	\centering
	\label{fig:SAN_scale}
\end{figure}

\noindent
\textbf{Partitioning.}
It is important for SAN to define scale partitioning for the selection of the sub-network of SAN and the reference scale for training of SAN.
The scale statistics for VOC PASCAL 2007 shows different medians and standard deviations of scales depending on the classes, but $231^2$ scale is the most common (Fig.~\ref{fig:SAN_scale}).
SAN is trained by using the convolutional features obtained by normalizing the RoI areas to the reference scale.
We conduct the experiment on three reference scales of $\{160^2, 224^2, 288^2\}$ around the common scale~$231^2$ and several partitions around the reference scale (Table~\ref{tab:table2}).
The best detection accuracy is obtained when the reference scale is $224^2$ and the scale space is partitioned into three sections: $(0^2,160^2]$, $(160^2,288^2)$, $[288^2,\infty)$.
The metric of MS COCO is defined over instance size: 
small~($Area<32^2$), medium~($32^2<Area<96^2$) and large~($96^2<Area$),
and
these partitions is doubled considering the process of resizing in detection network.
The best detection accuracy is obtained with
the partitions of $(0^2,64^2]$, $(64^2,192^2)$, $[192^2,\infty)$ at the reference scale~$128^2$
for MS COCO.
\\

\begin{table}[t]
	\caption{Varying partitions}
	\begin{center}
		\begin{tabular}{c|c|c|c}
			\toprule 
			$N_{partitions}$ & reference scale & partitioning & mAP(\%) \\
			\midrule 
			1 (R-FCN) 			& -  & - & 79.37 \\
			2 			& $160^2$ & $(0^2,160^2], [160^2,\infty)$ & 80.08 \\
			2 			& $224^2$ & $(0^2,224^2], [224^2,\infty)$ & 80.44 \\
			2 			& $288^2$ & $(0^2,288^2], [288^2,\infty)$ & 80.29 \\
			3 			& $224^2$ & $(0^2,160^2], (160^2,288^2), [288^2,\infty)$ & 80.57 \\
			3 			& $224^2$ & $(0^2,112^2], (112^2,448^2), [448^2,\infty)$ & 80.55 \\
			3 			& $224^2$ & $(0^2,112^2], (112^2,224^2),[224^2,\infty)$ & 80.16 \\
			\bottomrule 
		\end{tabular}
	\end{center}
	\label{tab:table2}
\end{table}

\begin{table}[t]
	\caption{Gaussian vs. Identity initialization}
	\begin{center}
		\begin{tabular}{c|c|c}
			\toprule 
			initialization method & $N_{partitions}$ & mAP(\%) \\
			\midrule 
			gaussian & 2& 79.85 \\
			gaussian & 3& 79.97 \\
			identity & 2& 80.44 \\
			identity & 3& 80.57 \\
			\bottomrule 
		\end{tabular}
	\end{center}
	\label{tab:table3}
\end{table}

\noindent
\textbf{Initialization.}
The initialization methods for the sub-networks belonging to SAN is an important issue.
Unlike ResNet-101 for the detection network, SAN does not have any pre-trained weights,
but we have the clue to the initialization: the role of SAN is reducing the difference between convolutional features for the scale difference.
Because SAN should output almost the same convolutional features in adjacent scales,
we need to initialize the weights to an identity matrix and biases to zero instead of the widely used initialization methods; gaussian, xavier~\cite{glorot2010understanding}, and MSRA~\cite{he2015delving}.
We compare this with the gaussian initialization and figure out that it is difficult to learn SAN without the initialization with an identity matrix (Table~\ref{tab:table3}).
\\

\noindent
\textbf{Pooling Method.}
Apart from the pooling method used in the detection frameworks, the pooling method is also needed for the learning of SAN.
We compare the average pooling (\texttt{AVE}) extracting the average value of a given area and the max pooling (\texttt{MAX}) extracting the maximum value of a given area (Table~\ref{tab:table4}).
In the case of R-FCN, SAN learned with the average pooling shows slightly higher detection accuracy than SAN learned with the max pooling.
\\

\noindent
\textbf{Mini-batch.}
To train SAN, we need the convolutional features for a scale normalized image at the reference scale.
However, since the additional convolutional operations are a time-consuming process, we selectively extract only a part of a mini-batch by normalizing it with a square of $224^2$.
We shows the effect of the number of samples on the detection accuracy and the best results are obtained with 16 samples (Table~\ref{tab:table5}).
\\

\begin{table}[b]
	\centering
	\begin{minipage}[t]{.45\linewidth}
		\centering
		\caption{Average vs Max pooling}
		\begin{center}
			\begin{tabular}{c|c|c}
				\toprule 
				$N_{partitions}$ & pooling method & mAP(\%) \\
				\midrule 
				2 & \texttt{AVE} & 80.44 \\
				3 & \texttt{AVE} & 80.57 \\
				3 & \texttt{MAX} & 80.35 \\
				\bottomrule 
			\end{tabular}
		\end{center}
		\label{tab:table4}
	\end{minipage}
	\centering
	\begin{minipage}[t]{.45\linewidth}
		\centering
		\caption{Varying mini-batch}
		\begin{center}
			\begin{tabular}{c|c|c}
				\toprule 
				$N_{partitions}$ & $N_{x}$ & mAP(\%) \\
				\midrule 
				2 & 16 & 80.44 \\
				3 & 4 & 80.37 \\
				3 & 8 & 80.57 \\
				3 & 16 & 80.57 \\
				\bottomrule 
			\end{tabular}
		\end{center}
		\label{tab:table5}
	\end{minipage} 
\end{table}

\begin{figure}[pt]
	
	\subfigure{\includegraphics[width=4cm]{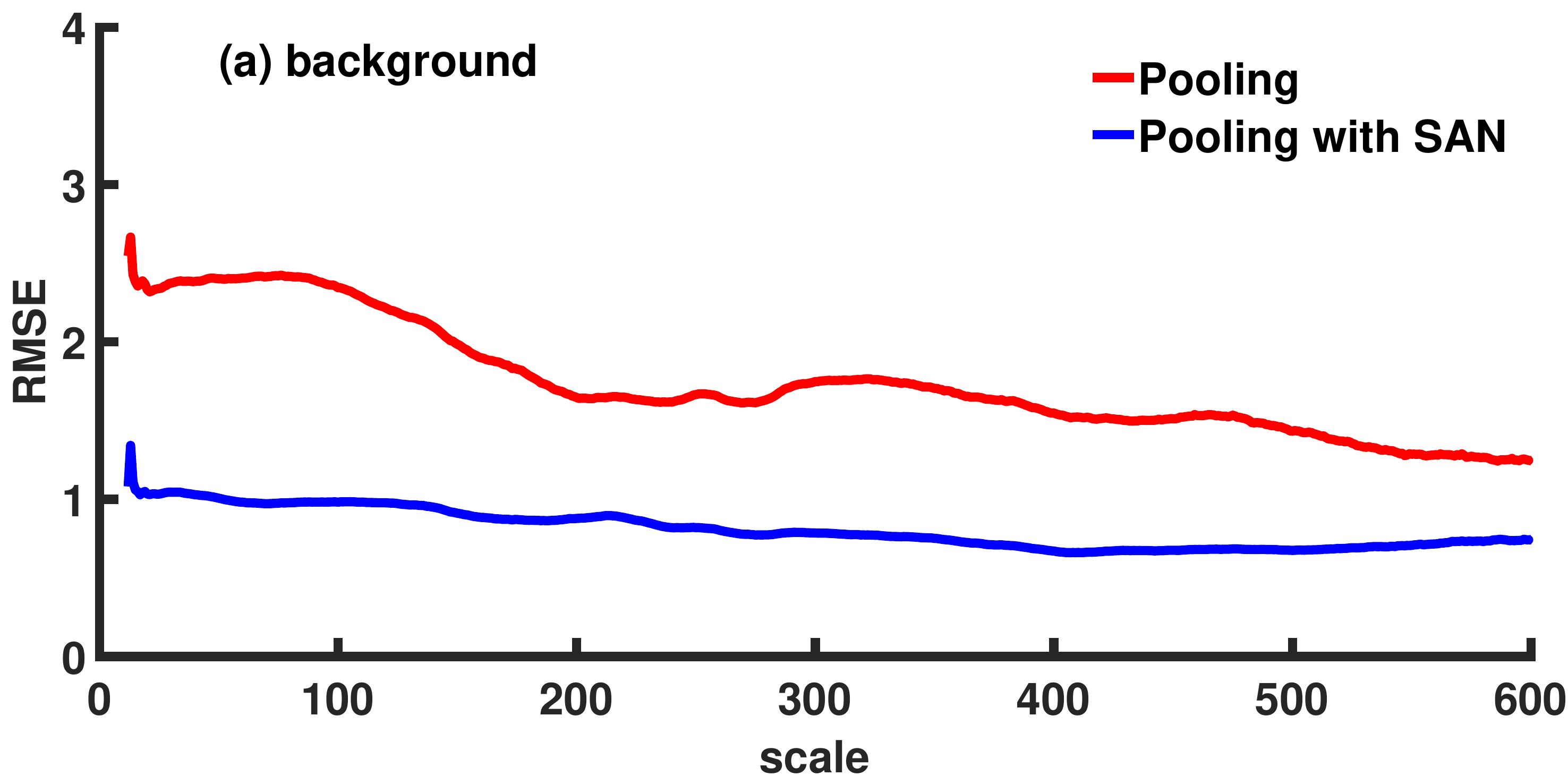}}
	\subfigure{\includegraphics[width=4cm]{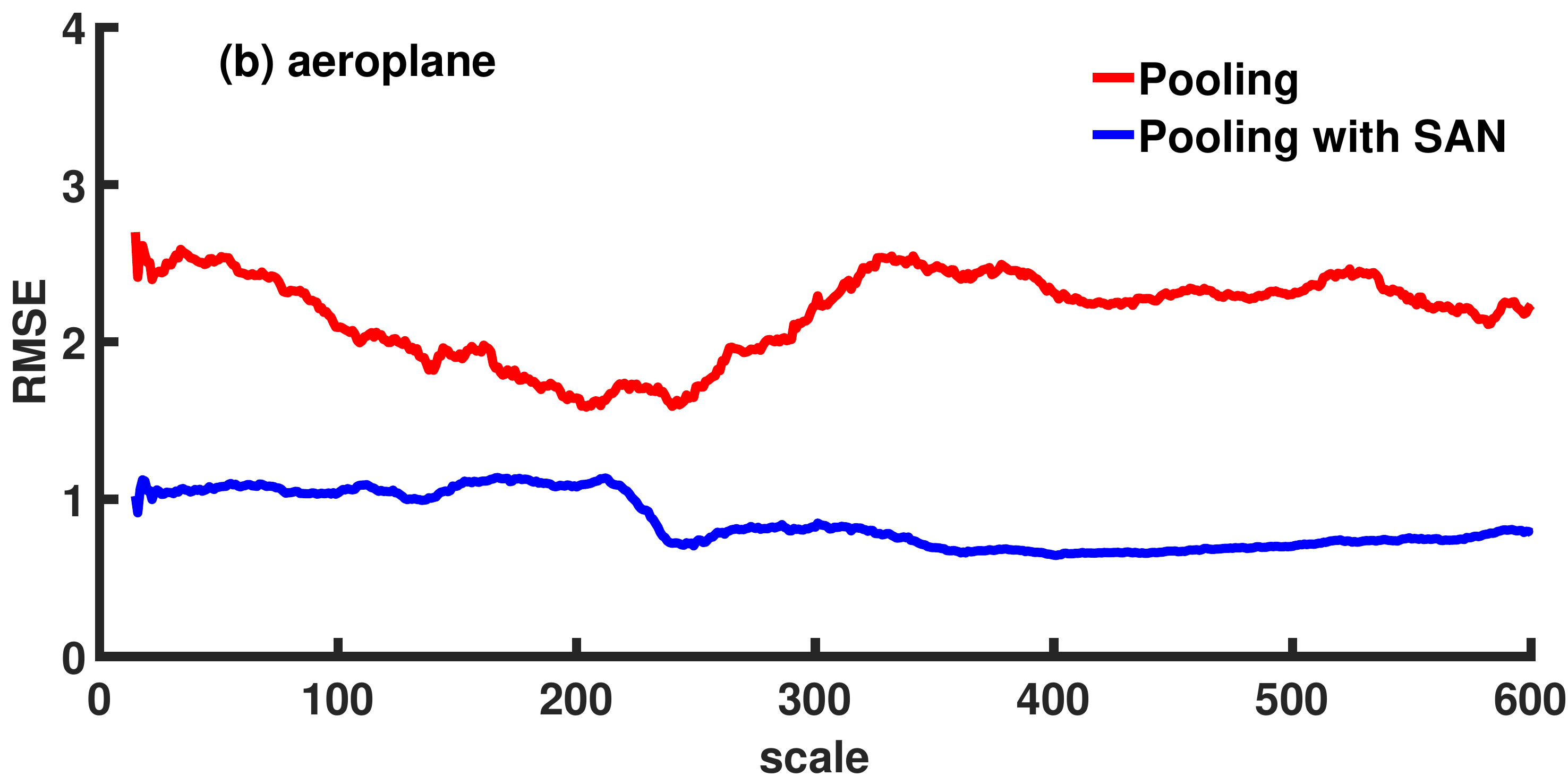}}
	\subfigure{\includegraphics[width=4cm]{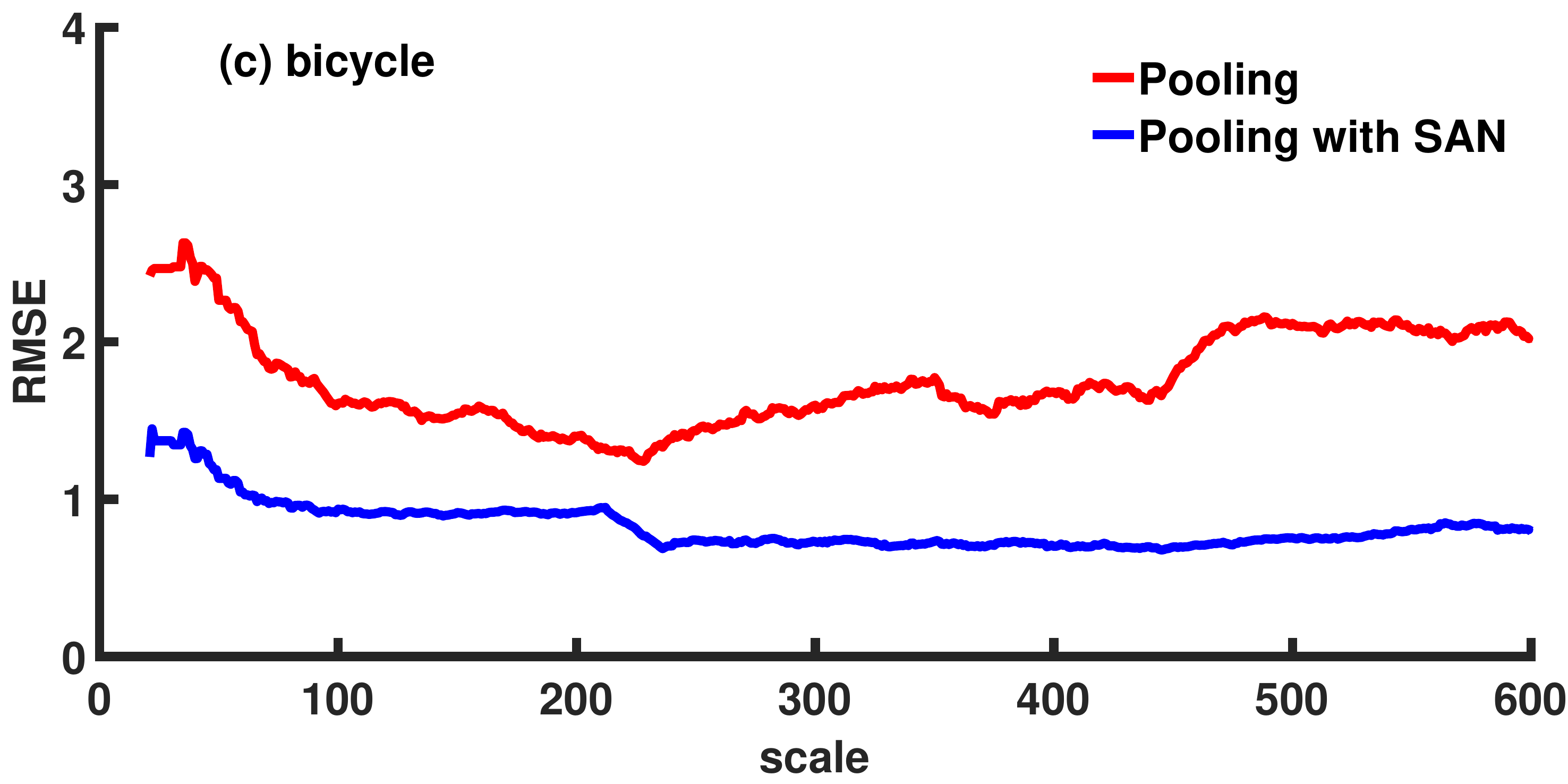}}
	\\
	\subfigure{\includegraphics[width=4cm]{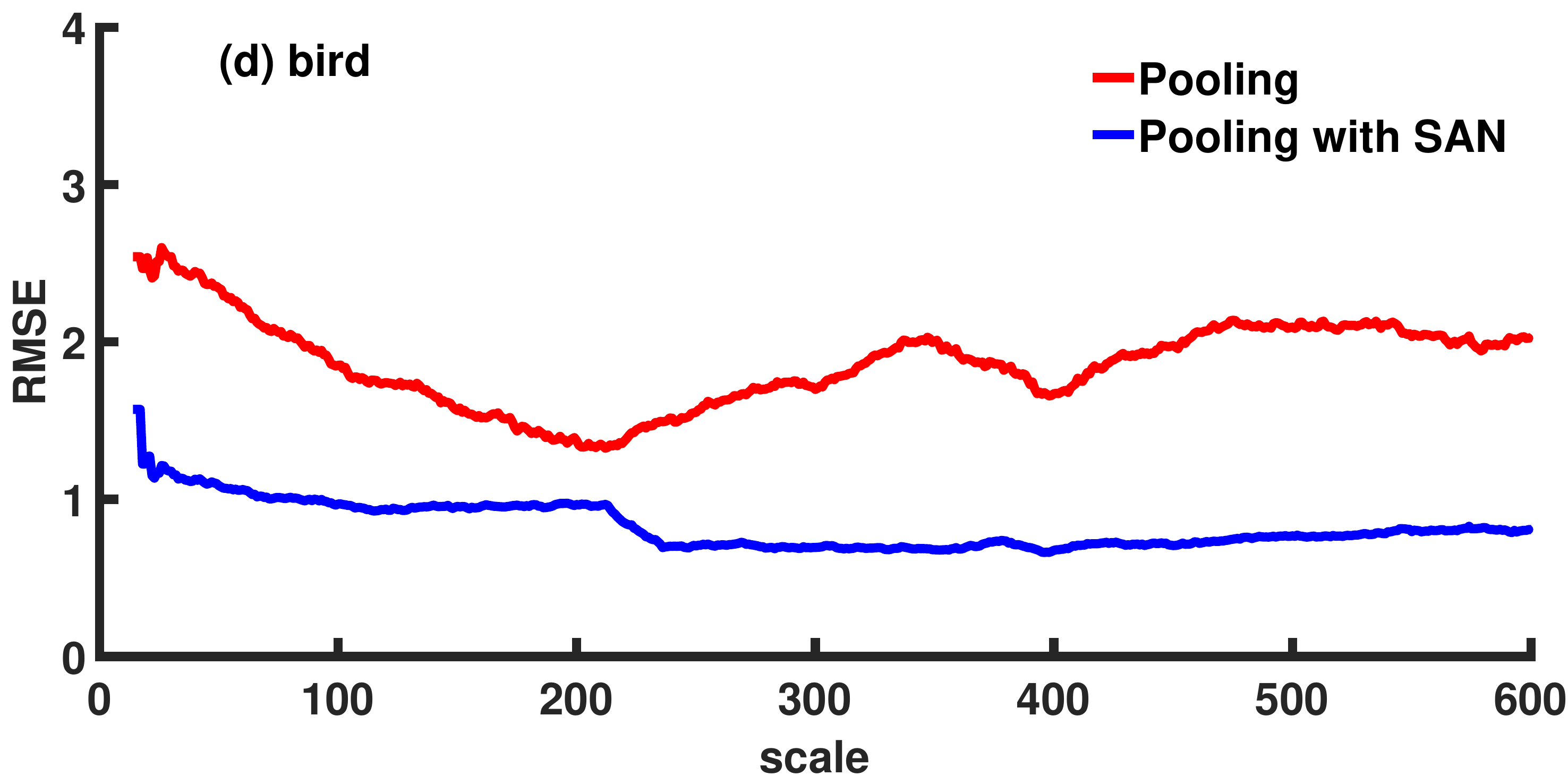}}
	\subfigure{\includegraphics[width=4cm]{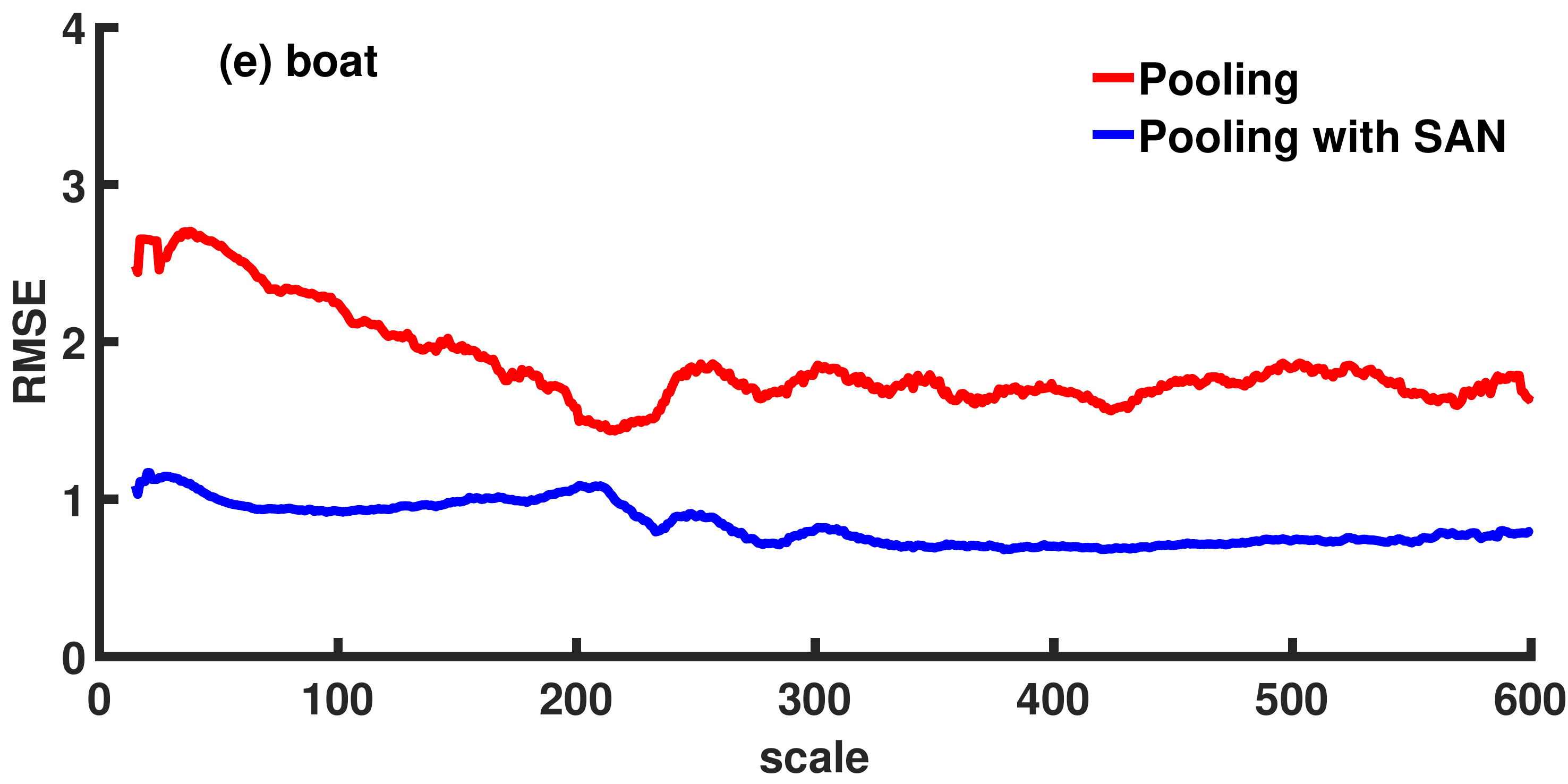}}
	\subfigure{\includegraphics[width=4cm]{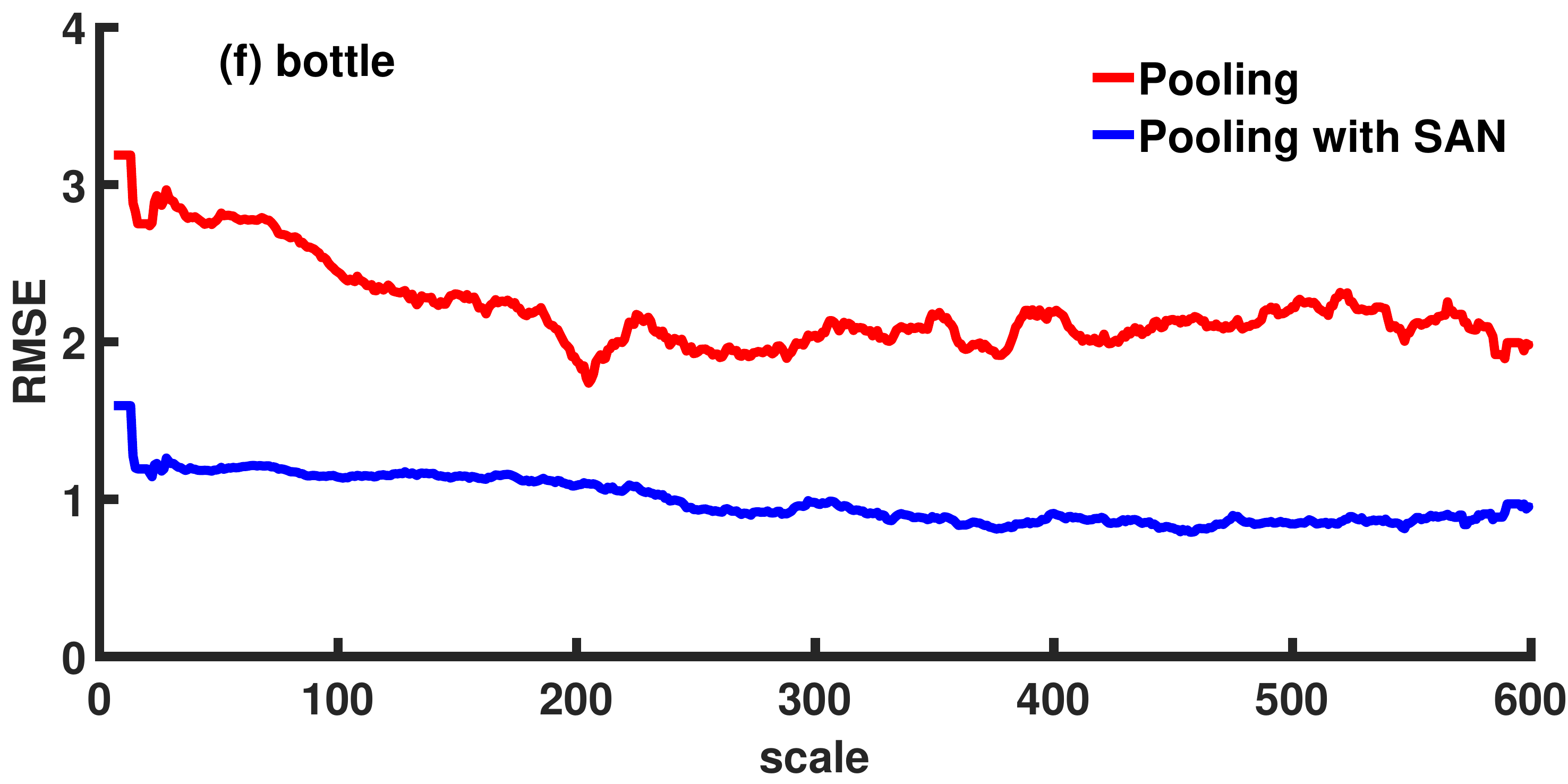}}
	\\
	\subfigure{\includegraphics[width=4cm]{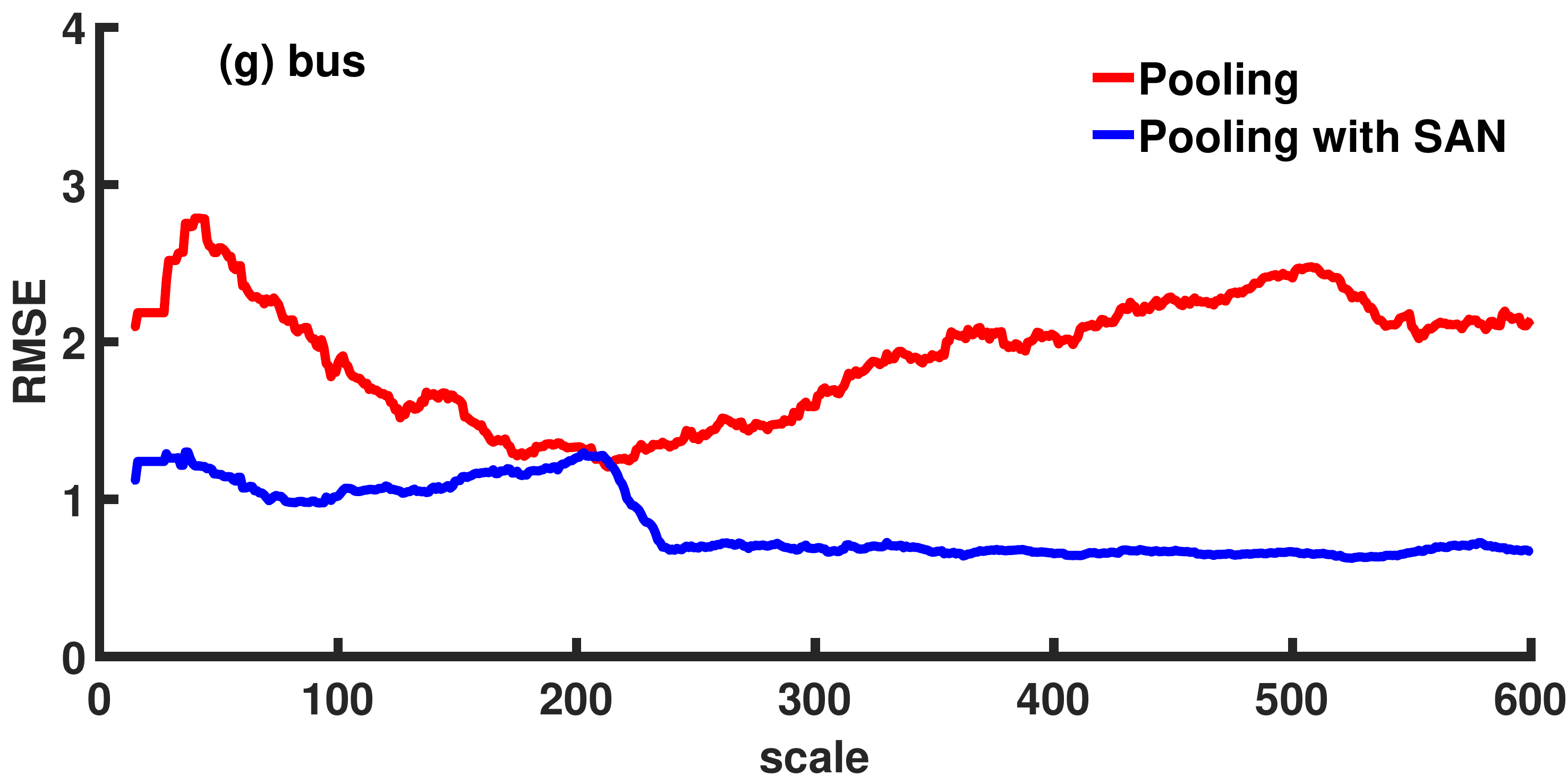}}
	\subfigure{\includegraphics[width=4cm]{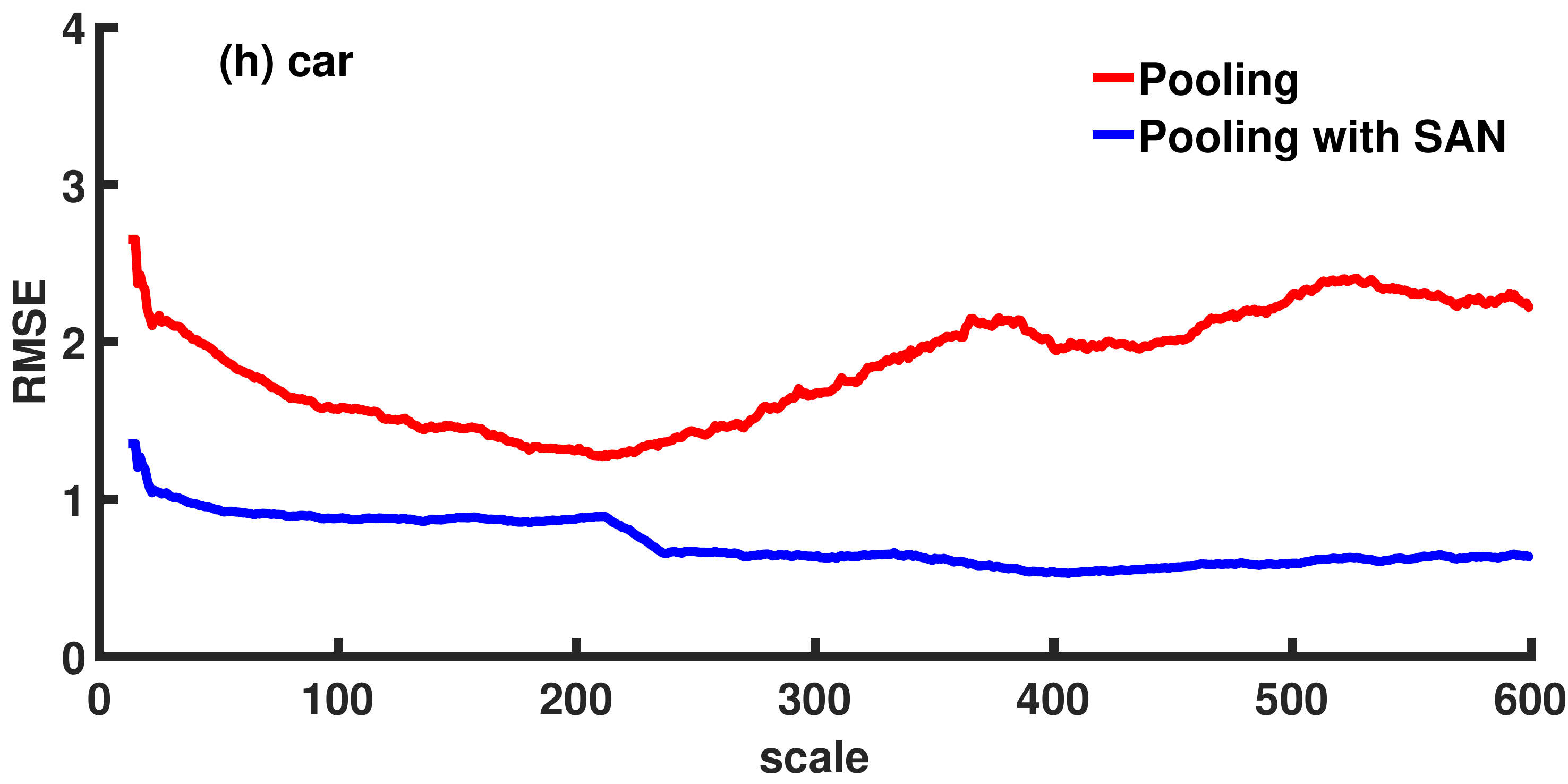}}
	\subfigure{\includegraphics[width=4cm]{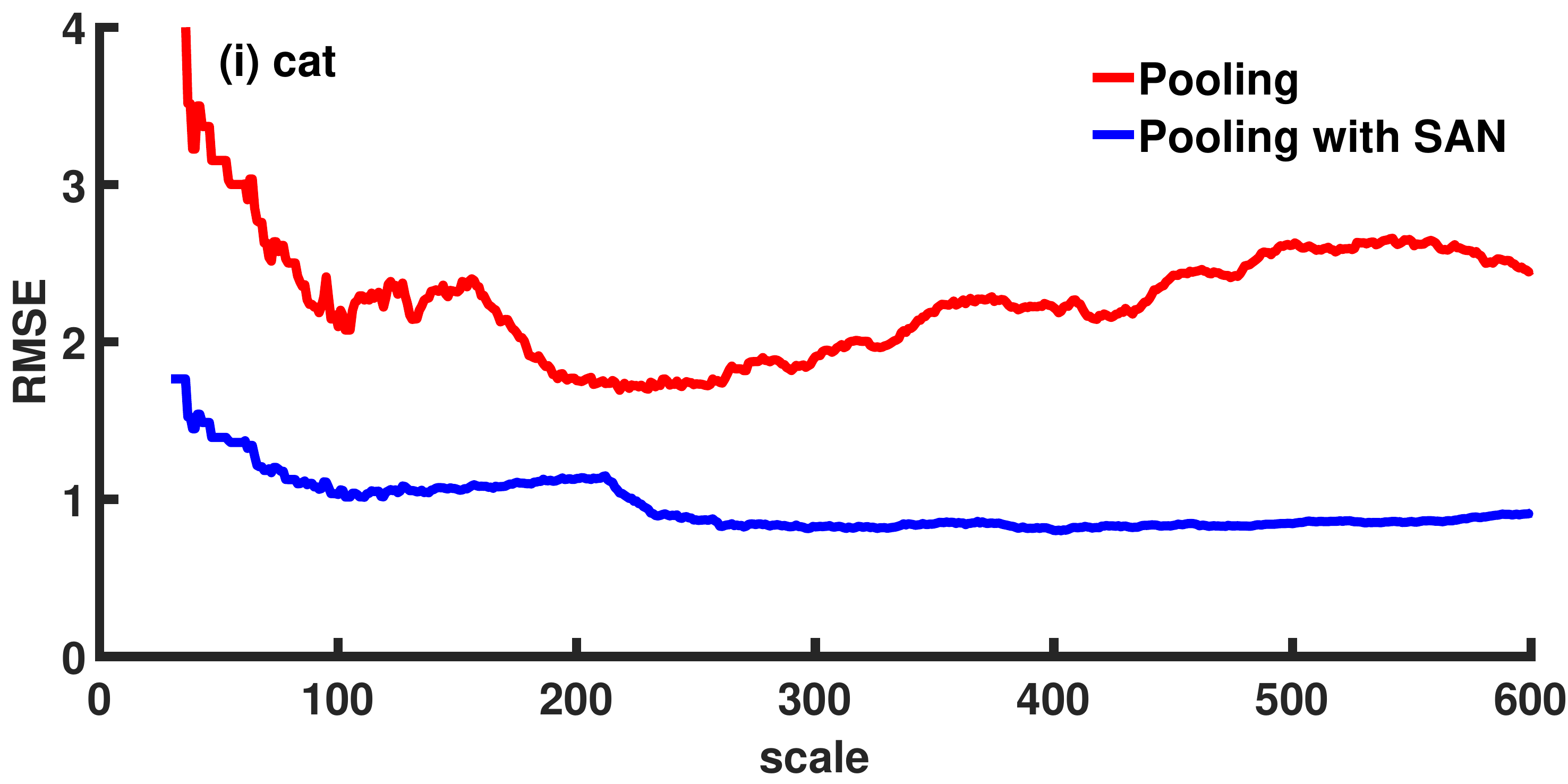}}
	\\
	\subfigure{\includegraphics[width=4cm]{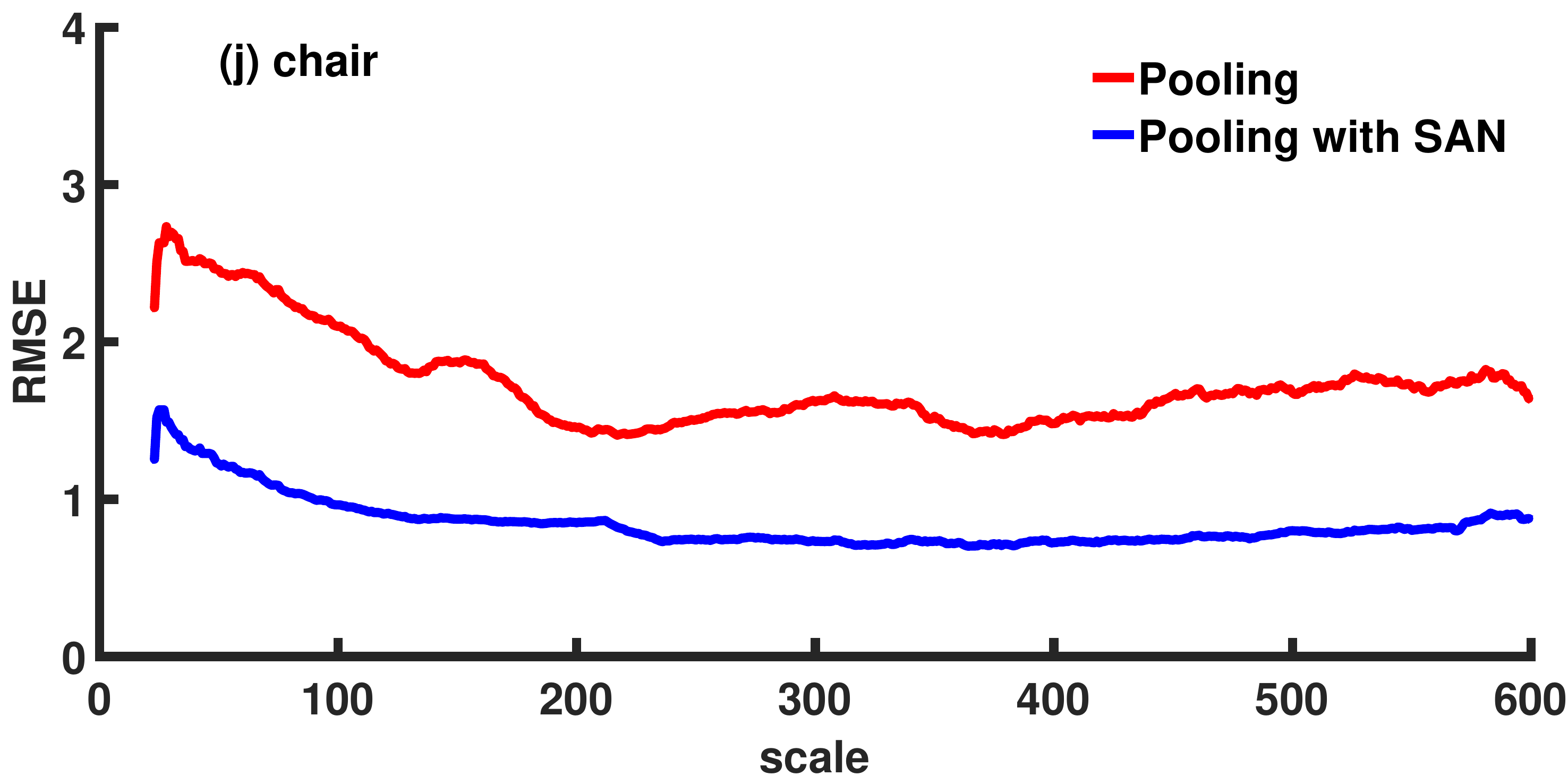}}
	\subfigure{\includegraphics[width=4cm]{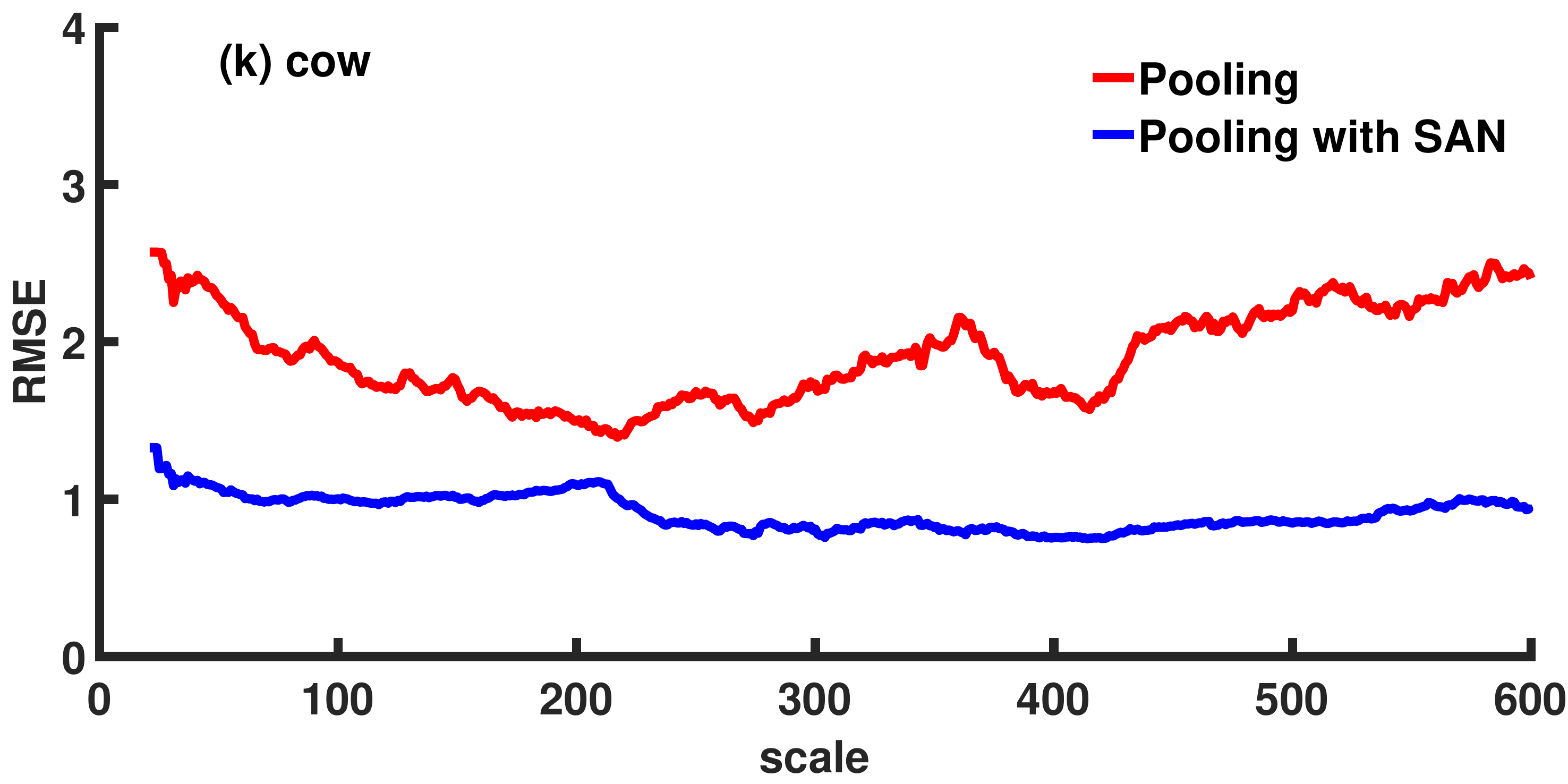}}
	\subfigure{\includegraphics[width=4cm]{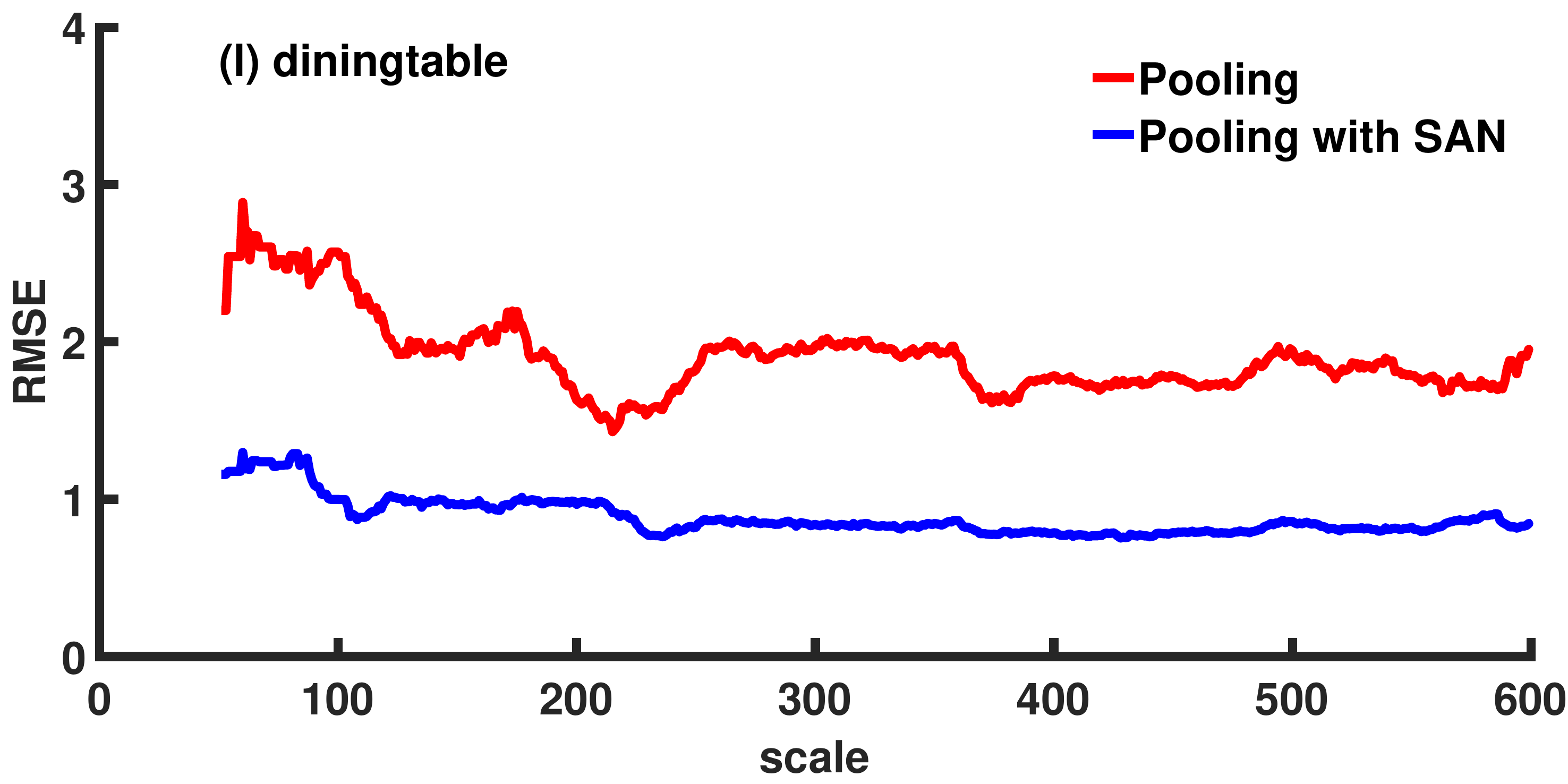}}
	\\
	\subfigure{\includegraphics[width=4cm]{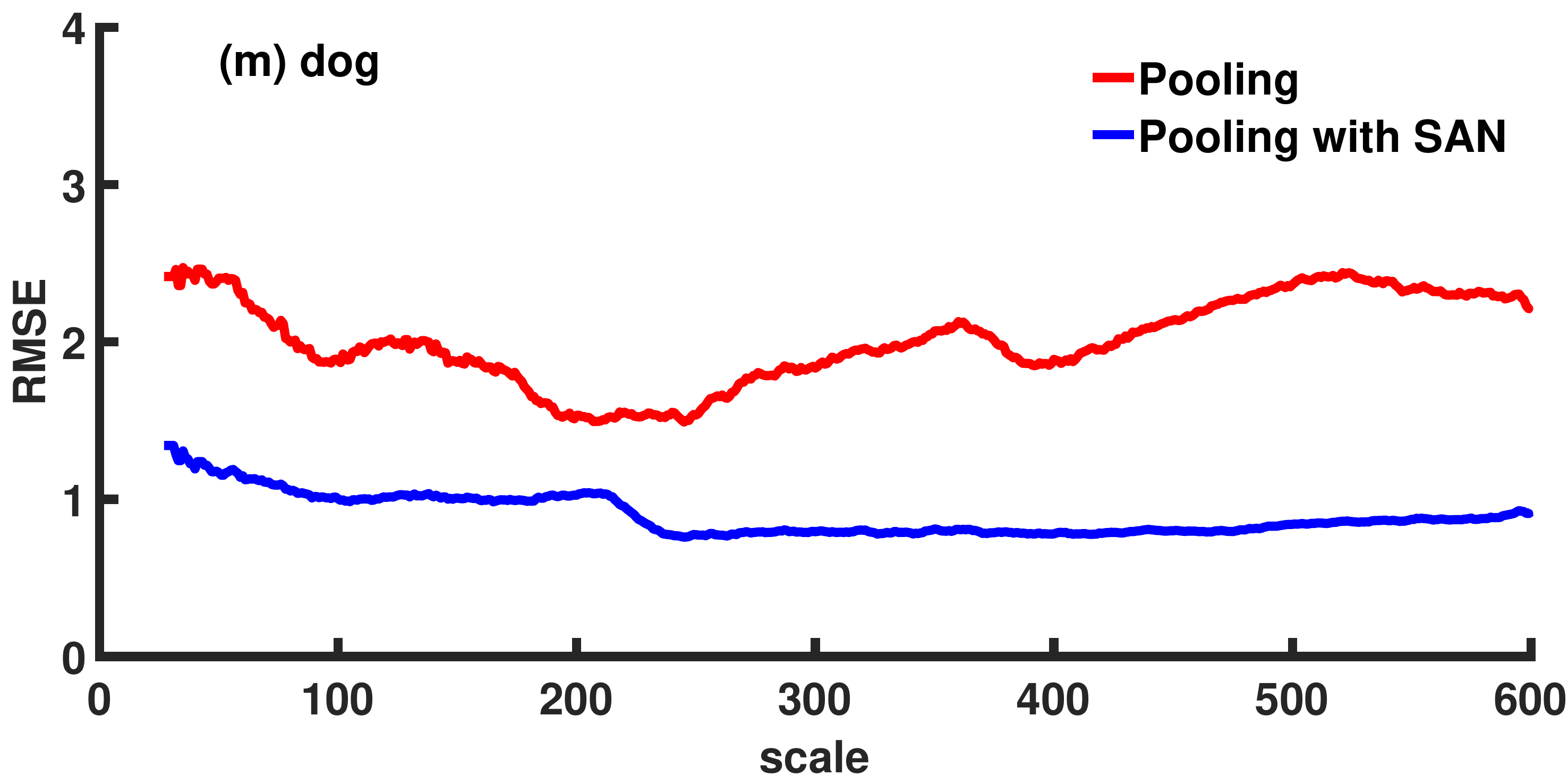}}
	\subfigure{\includegraphics[width=4cm]{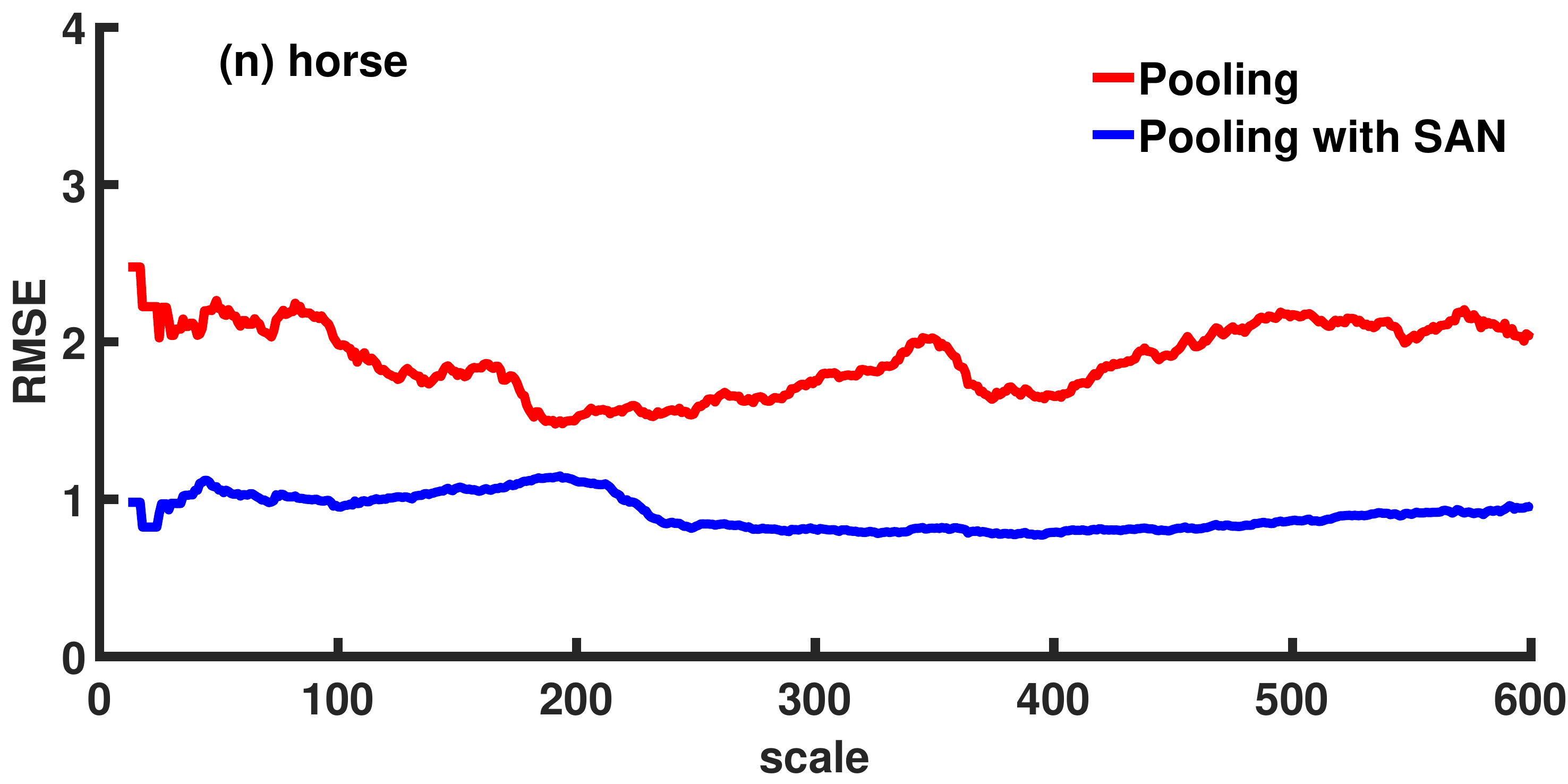}}
	\subfigure{\includegraphics[width=4cm]{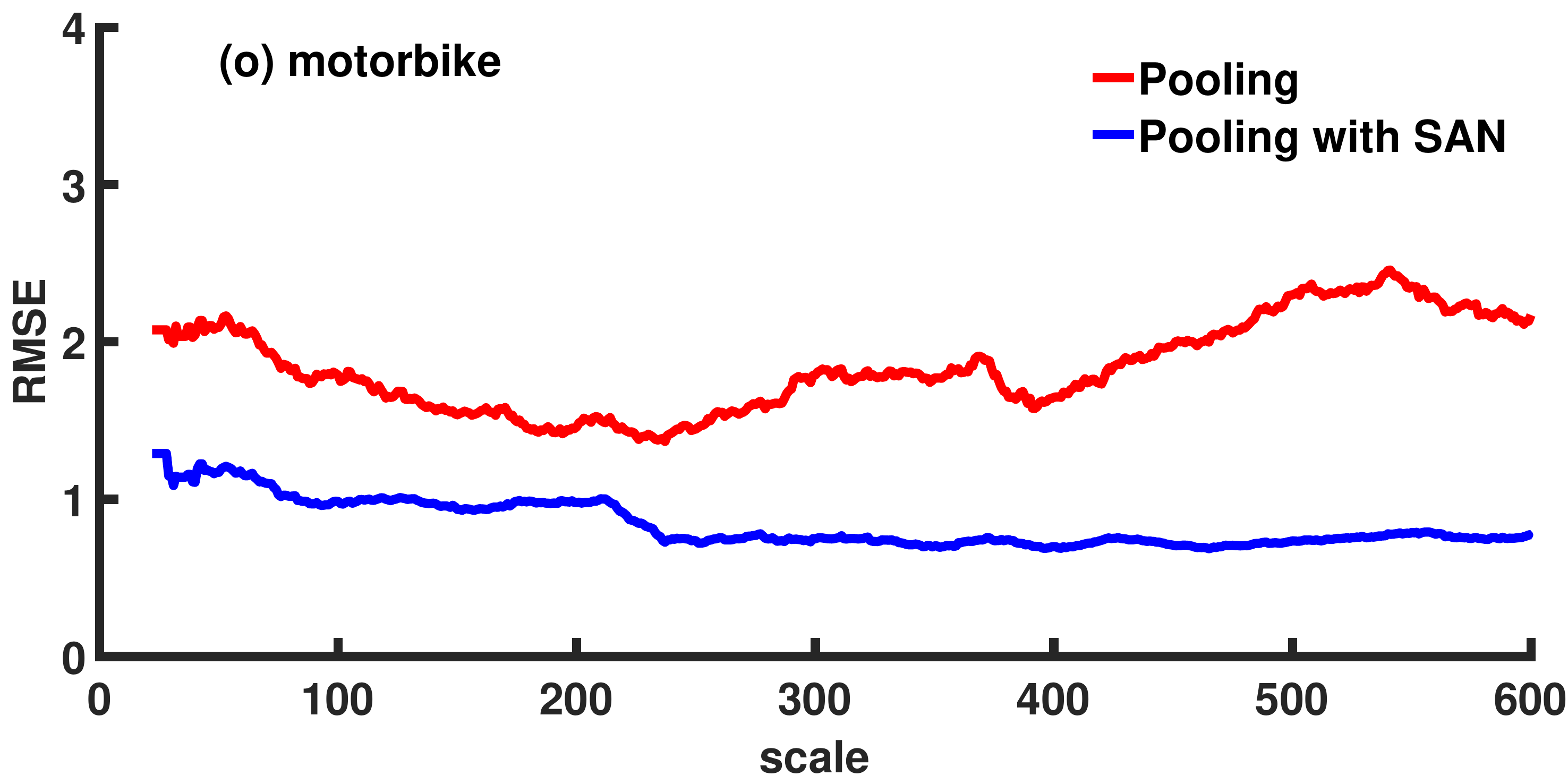}}
	\\
	\subfigure{\includegraphics[width=4cm]{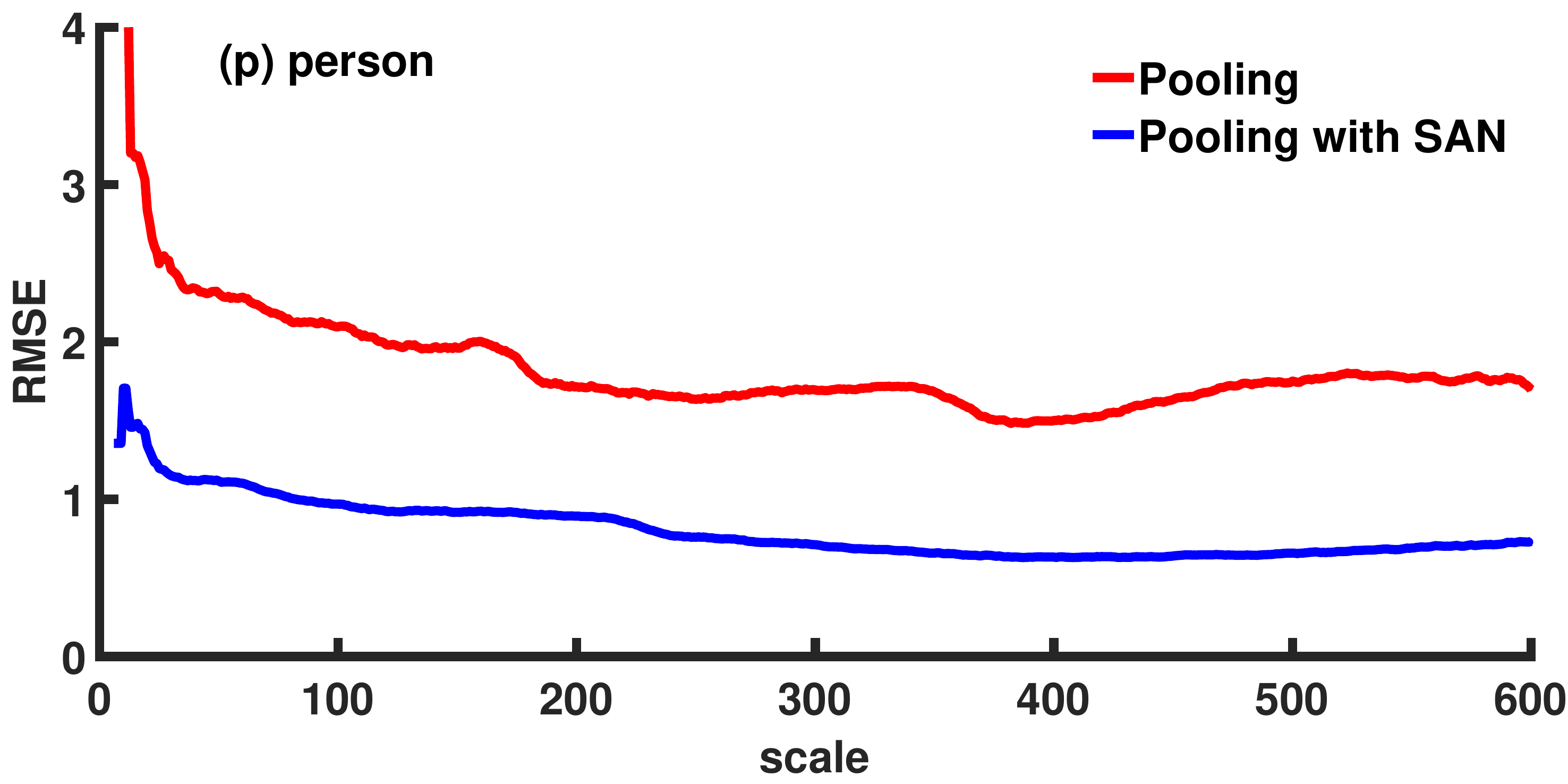}}
	\subfigure{\includegraphics[width=4cm]{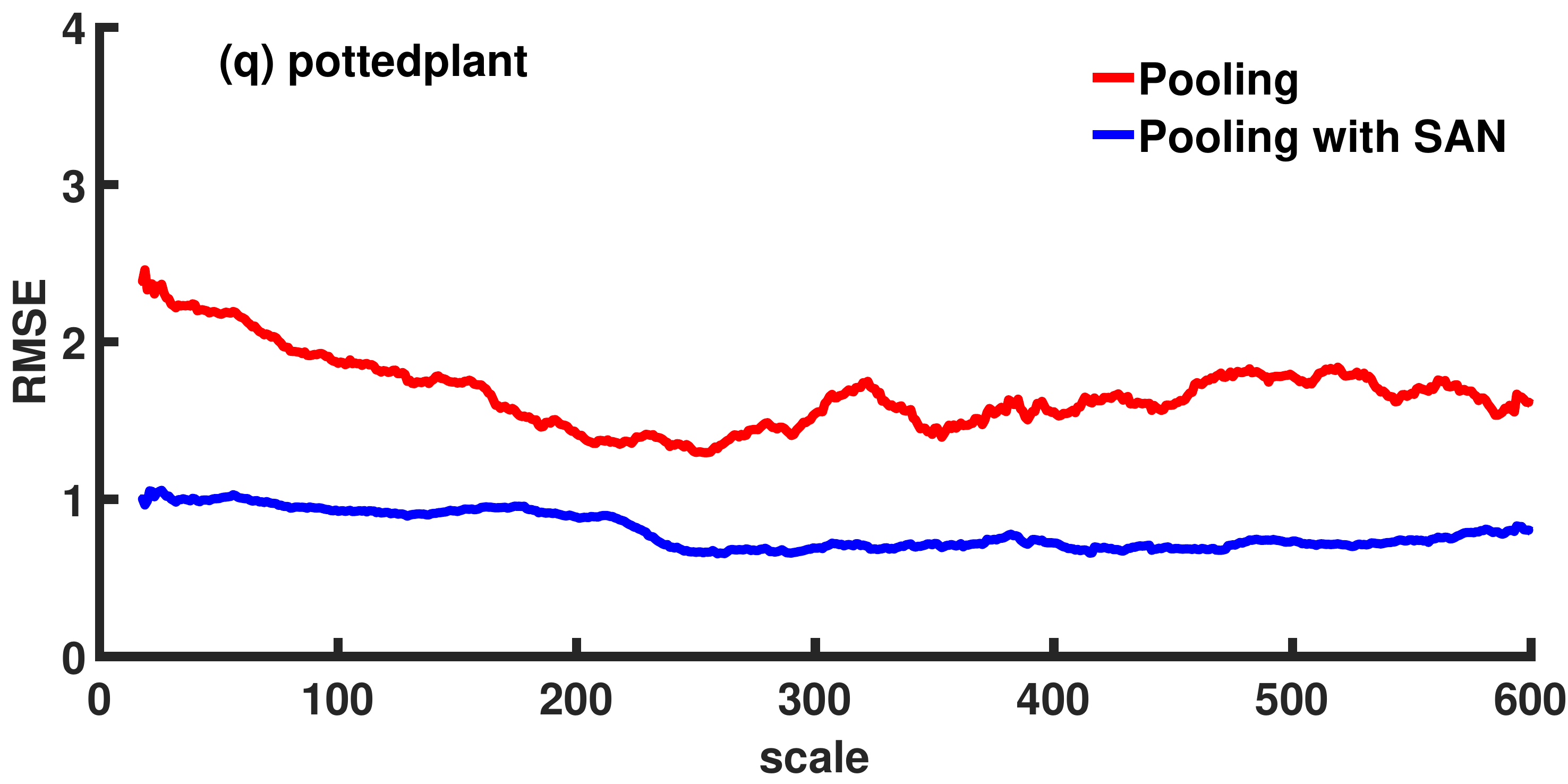}}
	\subfigure{\includegraphics[width=4cm]{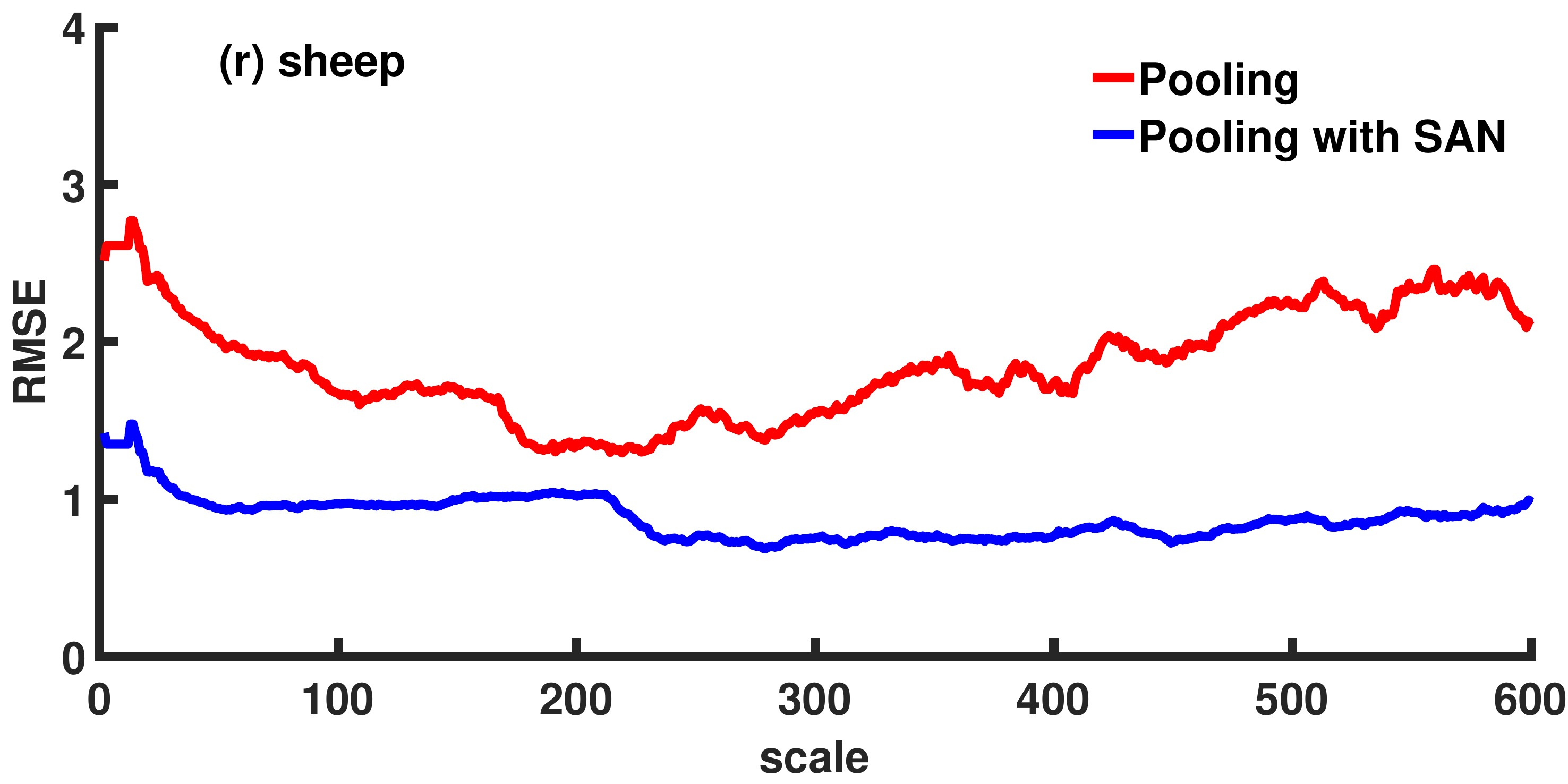}}
	\\
	\subfigure{\includegraphics[width=4cm]{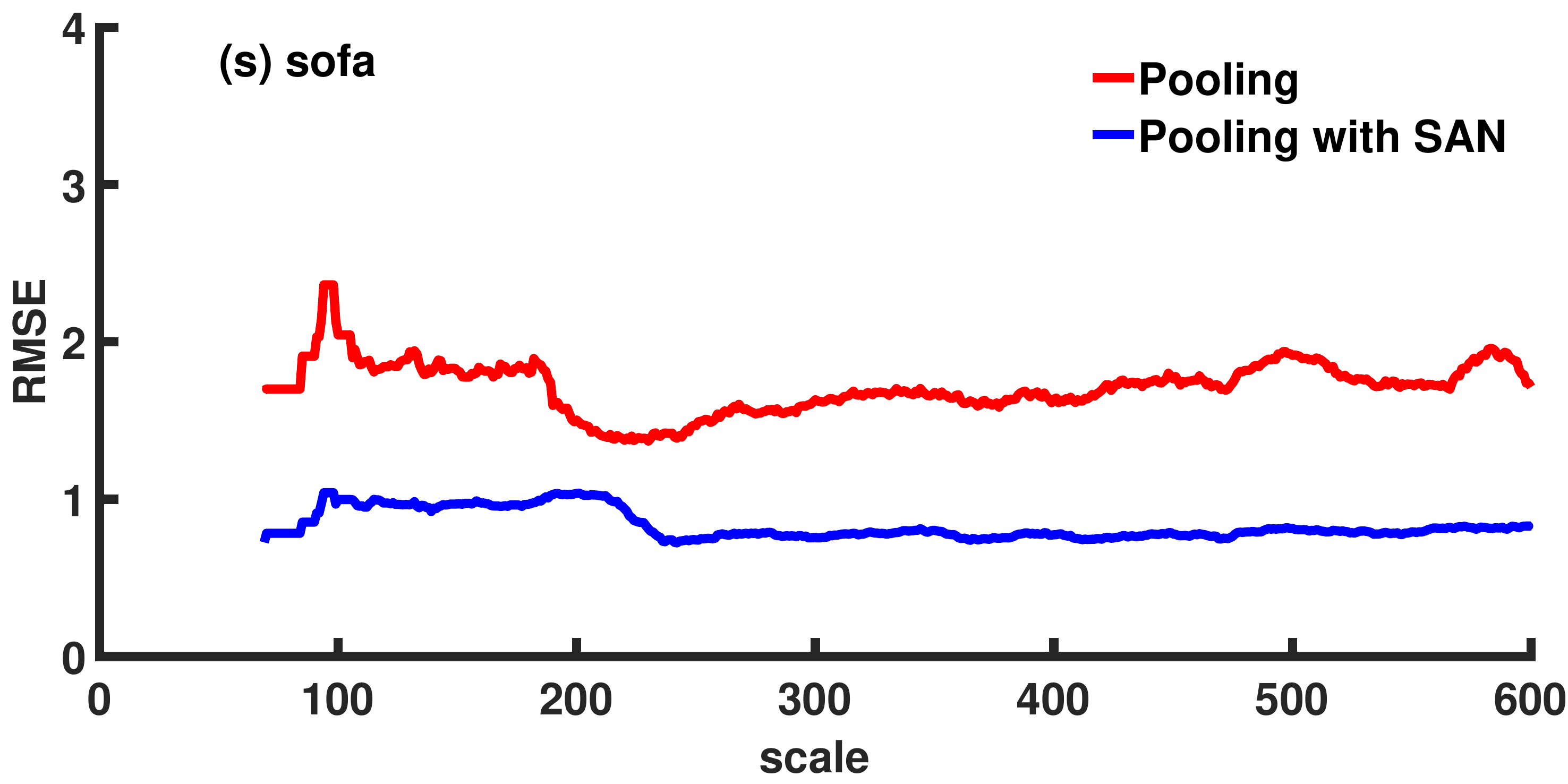}}
	\subfigure{\includegraphics[width=4cm]{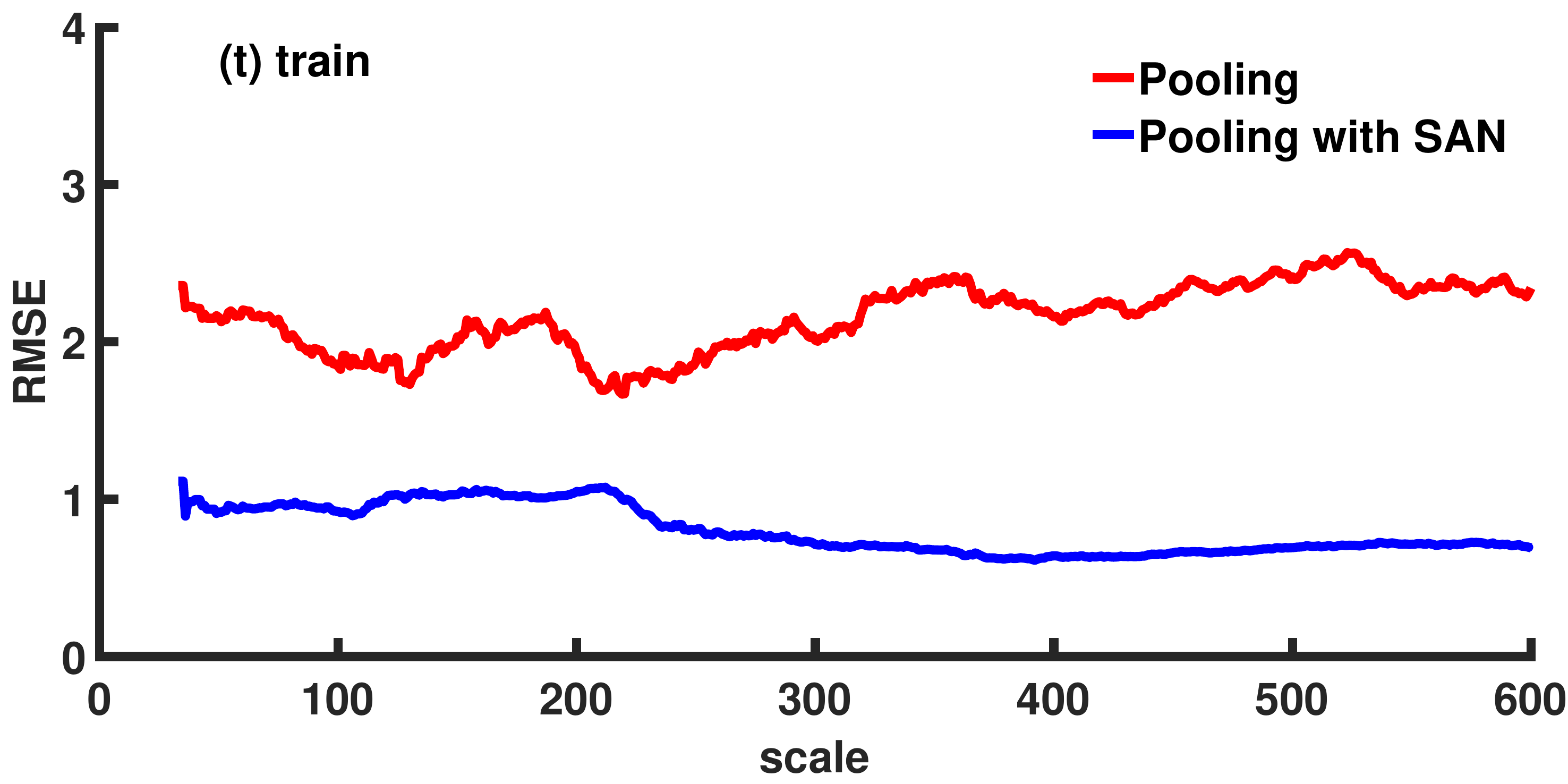}}
	\subfigure{\includegraphics[width=4cm]{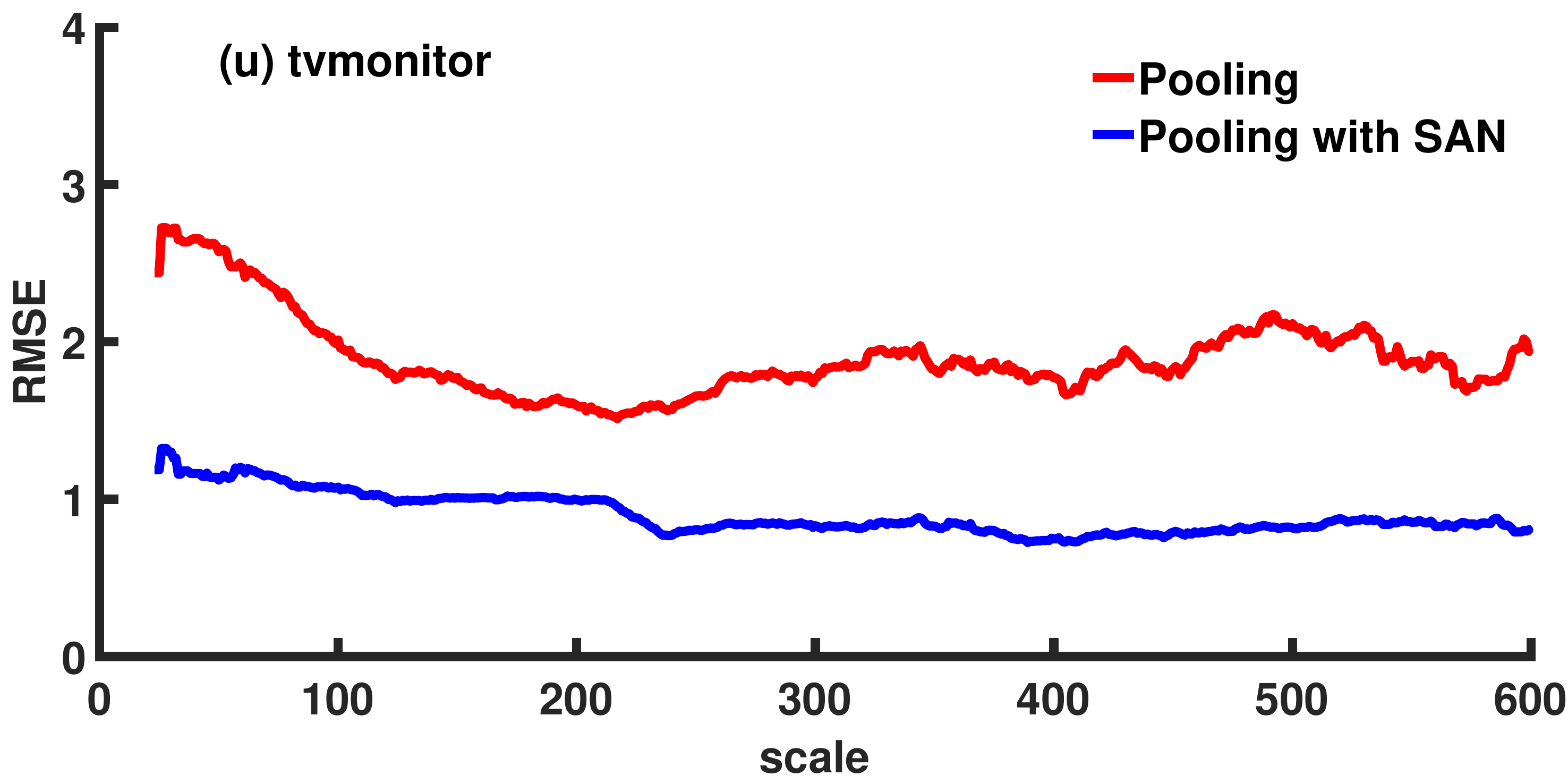}}
	\caption{The distribution for RMSE, which is a root mean squared error between the convolutional features extracted by RoI pooling with and without SAN, for 21 classes in VOC PASCAL}
	\centering
	\label{fig:sandist}
\end{figure}

\noindent
\textbf{Effectiveness of SAN.}
We define the convolutional feature at the reference scale as a scale-invariant feature and measure RMSE between the convolutional features~$z_{i,s}$ for sample $i$ at scale $s$ and the reference scale $s_0$, to demonstrate the effectiveness of SAN.
RMSEs for a convolutional feature without and with SAN are defined as 
\\
\begin{equation}
\begin{aligned}
\text{RMSE}(i,s|s_0) 		 &= \sqrt{ \frac{1}{N_c} \sum _ {\{x,y,c\} \in i}  \left[ z_{i,s}(x,y,c) - z_{i,s_0}(x,y,c)  \right] ^2 }, 
\end{aligned}
\end{equation}
\begin{equation}
\begin{aligned}
\text{RMSE}(i,s|s_0,f) &= \sqrt{ \frac{1}{N_c} \sum_ {\{x,y,c\} \in i}  \left[ f_{s} ( z_{i,s}(x,y,c) ) - z_{i,s_0}(x,y,c)  \right] ^2 }, 
\end{aligned}
\end{equation}
respectively, where $N_c$ is the number of channels and $f_{s}$ is the sub-network of SAN. 
The convolutional feature~$z_{i,s}$ is extracted using global average pooling to exclude the difference of spatial information.
We experimentally prove the validity of SAN by showing that SAN reduces RMSEs for all classes in VOC PASCAL (Fig.~\ref{fig:sandist}).

\section{Conclusion}
We propose a Scale Aware Network (SAN) that maps the convolutional features from the different scales onto a scale-invariant subspace to make CNN-based detection methods more robust to the scale variation, and also construct a unique learning method which considers purely the relationship between channels without the spatial information for the efficient learning of SAN.
To show the validity of our method, we visualize how convolutional features change according to the scale through a channel activation matrix and experimentally show that SAN reduces the feature differences in the scale space.
We evaluate our method on VOC PASCAL and MS COCO dataset.
Our method improves the mean Average Precision (mAP) by 1.2 points from R-FCN for VOC PASCAL 
and the COCO-style mAP by 2.1 point from Deformable R-FCN for MS COCO.
We demonstrate SAN for object detection by conducting several experiments on structures and parameters.	
The proposed SAN essentially improves the quality of convolutional features in the scale space, and can be generally applied to many CNN-based detection methods to enhance the detection accuracy with a slight increase in the computing time. 

As a future study, we will improve the performance by applying SAN to whole network including RPN, and try to study more deeply the relationship between convolutional features and scale normalization.
In addition, we plan to improve not only the object detection but also the general influence of the scale that can exist in many areas of computer vision.

\section*{Acknowledgement}
This work was supported by IITP grant funded by the Korea government (MSIT)
(IITP-2014-3-00059, Development of Predictive Visual Intelligence Technology, 
IITP-2017-0-00897, SW Starlab support program,
and 
IITP-2018-0-01290, Development of Open Informal Dataset and Dynamic Object Recognition Technology Affecting Autonomous Driving)

\begin{figure}[!pt]
	\begin{center}
		\subfigure{\includegraphics[width=12.2cm]{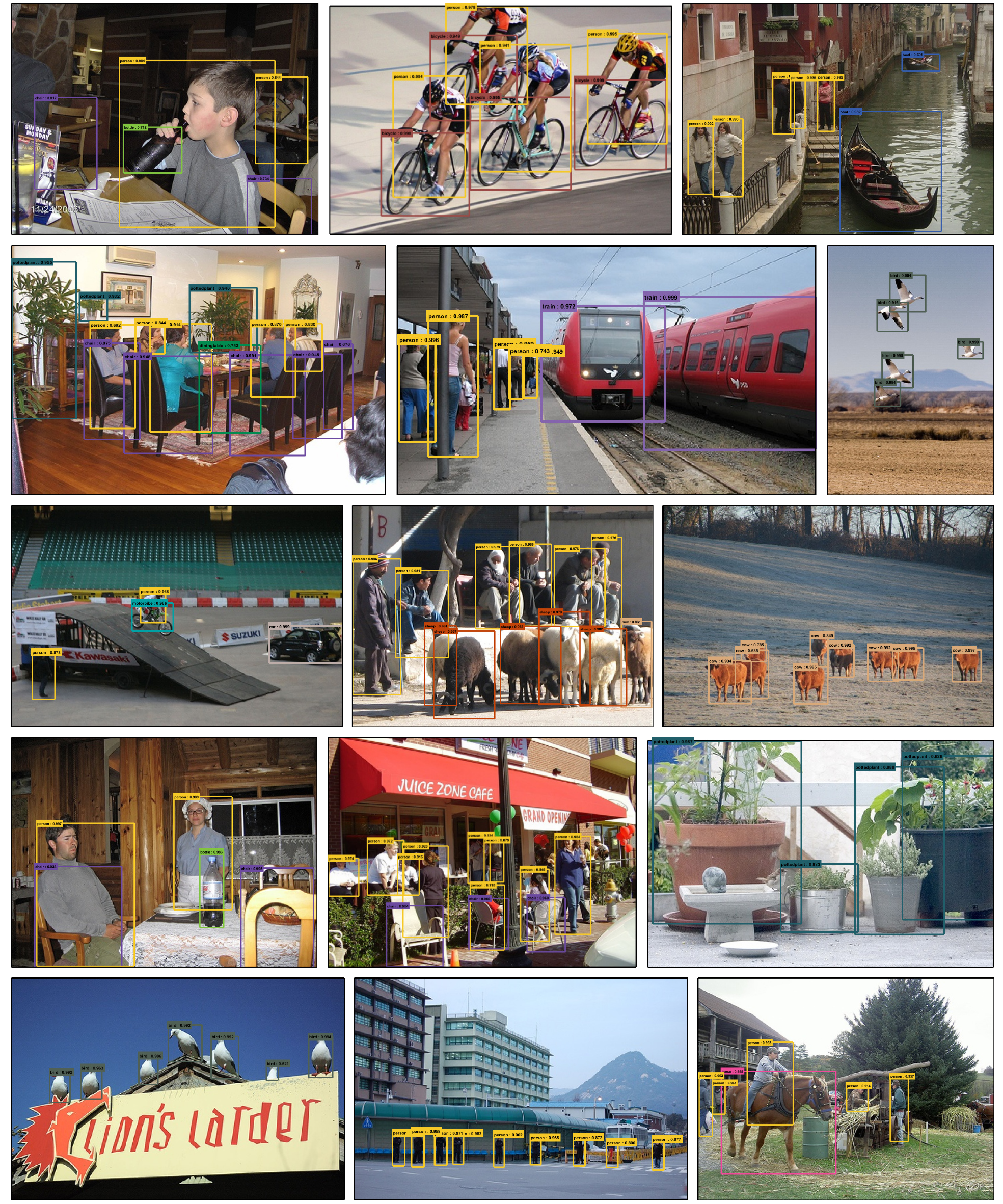}}
		\caption{Examples of object detection results on PASCAL VOC 2007 test set using R-FCN with SAN~(80.57\% mAP).
			The network is based on ResNet-101, and the training data is 07+12 trainval.
			A score threshold of 0.6 is used for displaying.
			The running time per image is 130 ms on NVidia Titan X Pascal GPU}
		\label{fig:RESULT}
	\end{center}
	
	\captionof{table}{Detailed detection results on PASCAL VOC 2007 test set}
	\begin{center}
		\scalebox{0.65}
		{
			\begin{tabular}
				{c|c|c| >{\centering}m{0.6cm}>{\centering}m{0.6cm}>{\centering}m{0.6cm}>{\centering}m{0.6cm}>{\centering}m{0.6cm}>{\centering}m{0.6cm}>{\centering}m{0.6cm}>{\centering}m{0.6cm}>{\centering}m{0.6cm}>{\centering}m{0.6cm}
					>{\centering}m{0.6cm}>{\centering}m{0.6cm}>{\centering}m{0.6cm}>{\centering}m{0.6cm}>{\centering}m{0.8cm}>{\centering}m{0.6cm}>{\centering}m{0.6cm}>{\centering}m{0.6cm}>{\centering}m{0.6cm}c}
				\toprule 
				Method & data & mAP
				& \scalebox{0.8}{aero} & \scalebox{0.8}{bike} & \scalebox{0.8}{bird} & \scalebox{0.8}{boat}
				& \scalebox{0.8}{bottle} & \scalebox{0.8}{bus} & \scalebox{0.8}{car} & \scalebox{0.8}{cat} & \scalebox{0.8}{chair} & \scalebox{0.8}{cow} 
				& \scalebox{0.8}{table} & \scalebox{0.8}{dog} & \scalebox{0.8}{horse} & \scalebox{0.8}{mbike} 
				& \scalebox{0.8}{person} & \scalebox{0.8}{plant} & \scalebox{0.8}{sheep} & \scalebox{0.8}{sofa} & \scalebox{0.8}{train} & \scalebox{0.8}{tv}  \\ 
				\midrule 
				R-FCN & 07+12 & \scalebox{0.8}{79.37}
				& \scalebox{0.8}{82.21} & \scalebox{0.8}{84.88} & \scalebox{0.8}{78.87} & \scalebox{0.8}{71.29} 
				& \scalebox{0.8}{68.67} & \scalebox{0.8}{88.54} & \scalebox{0.8}{87.10} & \scalebox{0.8}{89.11}
				& \scalebox{0.8}{67.86} & \scalebox{0.8}{87.06}  
				& \scalebox{0.8}{69.92} & \scalebox{0.8}{89.02} & \scalebox{0.8}{87.32} & \scalebox{0.8}{81.30} 
				& \scalebox{0.8}{79.73} & \scalebox{0.8}{52.16} & \scalebox{0.8}{78.17} & \scalebox{0.8}{80.91}
				& \scalebox{0.8}{83.43} & \scalebox{0.8}{79.88}  \\
				R-FCN(SAN) & 07+12 & \scalebox{0.8}{80.57}
				& \scalebox{0.8}{82.02} & \scalebox{0.8}{84.33} & \scalebox{0.8}{79.74} & \scalebox{0.8}{72.52} 
				& \scalebox{0.8}{70.16} & \scalebox{0.8}{87.33} & \scalebox{0.8}{87.72} & \scalebox{0.8}{89.45}
				& \scalebox{0.8}{68.74} & \scalebox{0.8}{87.51}  
				& \scalebox{0.8}{75.65} & \scalebox{0.8}{88.40} & \scalebox{0.8}{88.18} & \scalebox{0.8}{83.79} 
				& \scalebox{0.8}{81.05} & \scalebox{0.8}{53.71} & \scalebox{0.8}{81.75} & \scalebox{0.8}{81.04}
				& \scalebox{0.8}{87.21} & \scalebox{0.8}{81.13}  \\
				\bottomrule 
			\end{tabular}
		}
		\label{tab:table6}
	\end{center}
\end{figure}

%
%
\bibliographystyle{splncs04}
\bibliography{egbib}

\end{document}